\documentclass[10pt,journal,compsoc]{IEEEtran}

\usepackage{cite}
\usepackage{url}
\usepackage{ragged2e}
\usepackage{epsfig}
\usepackage{graphicx}
\usepackage{amsmath,amssymb} 
\usepackage{subfigure}
\usepackage{algorithm}
\usepackage{algorithmicx}
\usepackage{algpseudocode}
\usepackage{graphics}
\usepackage{threeparttable}
\usepackage{color}
\usepackage[normalem]{ulem}
\usepackage{multirow}
\usepackage{float}
\usepackage{amsfonts}
\usepackage{bm}
\usepackage{array}
\usepackage[table]{xcolor}
\usepackage{colortbl}
\usepackage{pifont}
\usepackage{diagbox}
\usepackage{rotating}
\usepackage{booktabs}
\usepackage{overpic}
\usepackage{textcomp}
\usepackage{contour}
\usepackage{enumitem}
\usepackage[colorlinks=true]{hyperref}

\RequirePackage{silence}
\def\ie{\emph{i.e.}}
\def\eg{\emph{e.g.}}
\def\etc{\emph{etc}}
\def\etal{{\em et al.~}}

\newcommand{\sArt}{state-of-the-art }

\definecolor{bblue}{rgb}{0,150,230}
\definecolor{mygray}{gray}{.92}

\newcommand{\figref}[1]{Fig.~\ref{#1}}
\newcommand{\tabref}[1]{Table~\ref{#1}}

\newcommand{\secref}[1]{Section \ref{#1}}

\newcommand{\tabincell}[2]{\begin{tabular}{@{}#1@{}}#2\end{tabular}}
\newcommand{\myPara}[1]{\vspace{10pt}\noindent$\bullet$~\textbf{#1} \quad}

\newcommand{\trb}[1]{\textbf{\textcolor{red}{#1}}}
\newcommand{\tgb}[1]{\textcolor{green}{#1}}
\newcommand{\tbb}[1]{\textcolor{blue}{#1}}
\newcommand{\supp}[1]{\textcolor{magenta}{#1}}

\def\ourdataset{CoSOD3k}
\def\ourmodel{\textit{CoEG-Net}}

\graphicspath{{./Imgs/}}
\DeclareGraphicsExtensions{.jpg,.pdf,.png}
\begin{document}

\title{Re-thinking Co-Salient Object Detection}

\author{Deng-Ping~Fan,
        Tengpeng Li,
        Zheng Lin,
        Ge-Peng Ji,
        Dingwen Zhang,
        Ming-Ming Cheng,~\IEEEmembership{Senior Member,~IEEE},
        Huazhu Fu,~\IEEEmembership{Senior Member,~IEEE},
        Jianbing Shen,~\IEEEmembership{Senior Member,~IEEE}\\
\IEEEcompsocitemizethanks{
\IEEEcompsocthanksitem D.-P.~Fan, Z.~Lin and M.-M.~Cheng are with the College of Computer Science, Nankai University, Tianjin, China.
(Email: dengpfan@gmail.com, frazer.linzheng@gmail.com, cmm@nankai.edu.cn)
\IEEEcompsocthanksitem T.~Li is with the B-DAT and CICAEET, Nanjing University
  of Information Science and Technology, Nanjing, China.
  (E-mail: ltpfor1225@gmail.com)
\IEEEcompsocthanksitem G.-P.~Ji is with the School of Computer Science,
  Wuhan University, Hubei, China.
  (E-mail: gepengai.ji@gmail.com)
\IEEEcompsocthanksitem D.~Zhang is with the Brain and Artificial Intelligence Laboratory, School of Automation, Northwestern Polytechnical University, Xi'an 710072, China.
  (E-mail: zhangdingwen2006yyy@gmail.com)
\IEEEcompsocthanksitem H.~Fu and J.~Shen are with the Inception Institute
  of Artificial Intelligence, Abu Dhabi, UAE.
  (E-mail: \{huazhu.fu, jianbing.shen\}@inceptioniai.org)
\IEEEcompsocthanksitem A preliminary version of this work has appeared in
  CVPR 2020~\cite{fan2020taking}.
\IEEEcompsocthanksitem Corresponding author: M.-M. Cheng.}
}

\markboth{IEEE TRANSACTIONS ON PATTERN ANALYSIS AND MACHINE INTELLIGENCE}%
{Fan \MakeLowercase{\textit{et al.}}}

\IEEEtitleabstractindextext{%
\begin{abstract}
  \justifying
  In this paper, we conduct a comprehensive study on the co-salient object
  detection (CoSOD) problem for images.
  CoSOD is an emerging and rapidly growing extension of salient object detection (SOD),
  which aims to detect the co-occurring salient objects in a group of images.
  However, existing CoSOD datasets often have a serious data bias,
  assuming that each group of images contains salient objects of similar
  visual appearances.
  This bias can lead to the ideal settings and effectiveness of models
  trained on existing datasets,
  being impaired in real-life situations,
  where similarities are usually semantic or conceptual.
  To tackle this issue, we first introduce a new benchmark,
  called \ourdataset~in the wild,
  which requires a large amount of semantic context,
  making it more challenging than existing CoSOD datasets.
  Our \ourdataset~consists of 3,316 high-quality,
  elaborately selected images divided into 160 groups with
  hierarchical annotations.
  The images span a wide range of categories, shapes, object sizes,
  and backgrounds.
  Second, we integrate the existing SOD techniques to build a unified,
  trainable CoSOD framework, which is long overdue in this field.
  Specifically, we propose a novel CoEG-Net that augments our prior model
  EGNet with a co-attention projection strategy to enable
  fast common information learning.
  CoEG-Net fully leverages previous large-scale SOD datasets and
  significantly improves the model scalability and stability.
  Third, we comprehensively summarize 40 cutting-edge algorithms,
  benchmarking 18 of them over three challenging CoSOD datasets
  (iCoSeg, CoSal2015, and our \ourdataset),
  and reporting more detailed (\ie, group-level) performance analysis.
  Finally, we discuss the challenges and future works of CoSOD.
  We hope that our study will give a strong boost to growth in the CoSOD
  community.
  The benchmark toolbox and results are available on our project page at
  \supp{\href{http://dpfan.net/CoSOD3K/}{http://dpfan.net/CoSOD3K/}}.
\end{abstract}


}

\maketitle

\IEEEdisplaynontitleabstractindextext

\IEEEpeerreviewmaketitle

\IEEEraisesectionheading{\section{Introduction}\label{sec:introduction}}

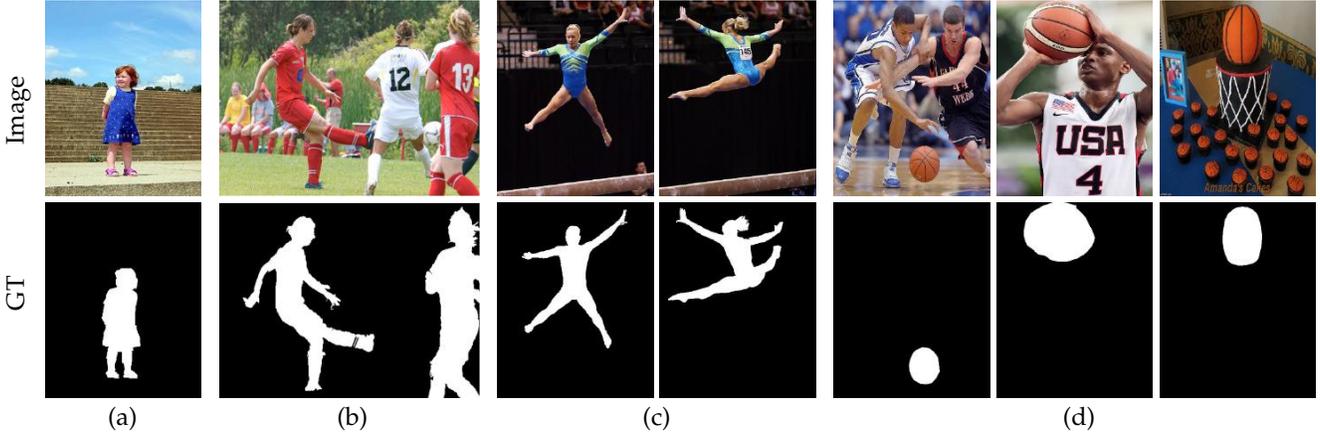
\begin{figure*}[t!]
	\centering
	\begin{overpic}[width=.93\textwidth]{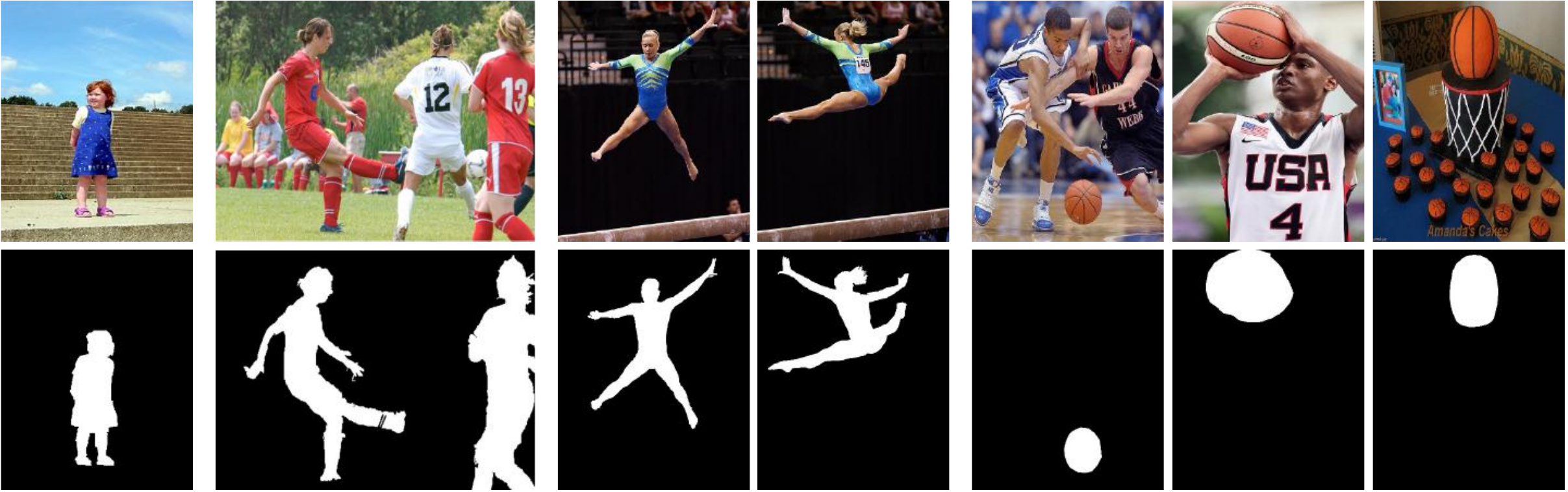}
	  \put(-3,20) {\rotatebox{90}{Image}}
    \put(-3,7){\rotatebox{90}{GT}}
    \put(5,-2) {(a)}
    \put(23,-2) {(b)}
    \put(47,-2) {(c)}
    \put(80,-2) {(d)}
  \end{overpic}
	\caption{Different salient object detection (SOD) tasks.
    (a) Traditional SOD~\cite{wang2017learning}.
    (b) Within-image co-salient object detection (CoSOD)~\cite{yu2018co},
    where common salient objects are detected from a single image.
    (c) Existing CoSOD, where salient objects are detected across
    a pair~\cite{li2011co} or a group~\cite{winn2005object} of images
    with similar appearances.
    (d) The proposed CoSOD in the wild,
    which requires a large amount of semantic context,
    making it more challenging than existing CoSOD.
  }\label{fig:taskExample}
\end{figure*}

\IEEEPARstart{S}{alient} object detection (SOD) in color images
\cite{fan2018salient,zeng2019towards,li2019nested,BorjiCVM2019,qin2021boundary},
RGB-D images~\cite{zhao2019contrast,fan2019rethinking,zhang2020uncertainty,fu2020siamese,zhou2020rgb},
and videos~\cite{fan2019shifting,Cong2019TIP,wang2020revisiting}
has been an active field of research in the computer vision community over
the past~\cite{hou2007saliency,li2016deep,zhang2017amulet,cong2018review,
Wang2019survey,su2019selectivity,bi2020hbox,ren2020co}.
SOD mimics the human vision system to detect the most
attention-grabbing object(s) in a single image,
as shown in \figref{fig:taskExample} (a).
As a extension of this, co-salient object detection (CoSOD) emerged recently
to employ a set of images.
The goal of CoSOD is to extract the salient object(s) that are common within
a single image
(\eg, red-clothed football players in \figref{fig:taskExample} (b))
or across multiple images (\eg, the blue-clothed gymnast in
\figref{fig:taskExample} (c)).
Two important characteristics of co-salient objects are local saliency
and global similarity.
Due to its useful potential, CoSOD has been attracting growing attention
in many applications,
including collection-aware crops~\cite{jacobs2010cosaliency},
co-segmentation~\cite{wang2016higher,Fu2015VideoCoseg},
weakly supervised learning~\cite{zhang2019capsal},
image retrieval~\cite{liu2013model,cheng2014salientshape},
and video foreground detection~\cite{fu2013cluster}.

As such, the CoSOD task has been rapidly growing in recent few years
\cite{zhang2018review,cong2018review},
with hundreds of related publications since 2010\footnote{Some representative
works can be found on \url{https://hzfu.github.io/proj_cosal_review.html}.}.
Most CoSOD datasets tend to focus on the appearance-similarity between objects
to identify the co-salient object across multiple images.
However, this leads to \emph{data selection bias}
\cite{fan2018salient,torralba2011unbiased} and is not always appropriate,
since, in real-world applications, the salient objects in a group of images
often vary in terms of \emph{texture}, \emph{scene}, and \emph{background}
(see our \ourdataset~dataset in \figref{fig:taskExample} (d)),
even if they belong to the same category.
%
%
In addition to the data selection bias,
CoSOD methods also suffer from two main limitations:

%
%

\textbf{(A) Completeness.}
 $\epsilon$ (Mean Absolute Error)~\cite{Cheng_Saliency13iccv} and
F-measure~\cite{achanta2009frequency} are two widely used metrics in
CoSOD/SOD model evaluation.
As discussed in~\cite{margolin2014evaluate},
these metrics have their inherent limitations.
To provide thorough and reliable conclusions,
we need introduce more accurate metrics \eg,
structural based evaluation metric or perceptual based evaluation metric.

\textbf{(B) Fairness.}
To evaluate the F-measure, the first step is to binarize a saliency map
into a set of foreground maps using different threshold values.
There are many binarization strategies~\cite{borji2015salient},
such as adaptive threshold, fixed threshold and so on.
However, different strategies will result in different F-measure performances.
Further, few previous works provide details on their binarization strategy,
leading to inconsistent F-measures for different researchers.

To address the aforementioned limitations,
we argue that integrating various publicly available CoSOD algorithms,
datasets, and metrics, and then providing a complete, unified benchmark,
is highly desired.
As such, we make four distinct contributions in this work:

\begin{itemize}
\item \textbf{First, we construct a challenging \ourdataset~dataset,
with more realistic settings.}
Our \ourdataset \footnote{Collecting the CoSOD dataset is more difficult
than the SOD dataset, that is why the previous largest CoSOD dataset,
\ie,~\cite{zhang2015co},
in the past 15 years  has only 2K images. Even for our 3K dataset, we
have spent 1 year to collect such high-quality dataset. Moreover,
we also pay more attention to provide high-quality hierarchical
annotations (\eg, image-level and object-/instance-level) to promote
related vision tasks rather than the size of the dataset.}
is the largest CoSOD dataset to date, with two aspects:
1) it contains 13 super-classes, 160 groups, and 3,316 images in total,
where each super-class is carefully selected to cover diverse scenes;
\eg, \emph{Vehicle}, \emph{Food}, \emph{Tool}, \etc.;
2) each image is accompanied by hierarchical annotations,
including category, bounding box, object, and instance,
which could greatly benefit various vision tasks (\eg, object proposal,
co-location, co-segmentation, co-instance detection, \etc.), as shown in
\figref{fig:CoSal3kExample}.

\item \textbf{Second, we present the first large-scale co-salient
object detection study,}
reviewing 40 state-of-the-art (SOTA) models, and
evaluating 18 of them on three challenging,
large-scale CoSOD datasets (iCoSeg, CoSal2015, and the proposed \ourdataset).
A convenient benchmark toolbox is also provided to integrate various
publicly available CoSOD datasets and multiple metrics for better
performance evaluation.
The benchmark toolbox and results have been made publicly available at
\supp{\href{https://dpfan.net/CoSOD3K/}{https://dpfan.net/CoSOD3K/}}.

\item \textbf{Third, we propose a simple but effective \ourmodel~baseline
for CoSOD,}
which uniformly and simultaneously embeds the appearance and semantic features
through a co-attention projection and a basic SOD network.
Comprehensive benchmarking results show that \ourmodel~outperforms the 18 SOTA
models.
Moreover, it also yields competitive visual results,
making it an efficient solution for the CoSOD task.

\vspace{5pt}
\item
\textbf{Finally, we make several interesting observations,
discuss the important issues arising from the benchmark results,
and suggest some future directions.}
Our study serves as a potential catalyst for promoting large-scale model
comparison for future CoSOD research.
\end{itemize}

\begin{table}
  \centering
  \footnotesize
  \begin{threeparttable}[thp!]
  \renewcommand{\arraystretch}{1.2}
  \setlength\tabcolsep{0.7pt}
  \caption{Statistics of existing CoSOD datasets and the proposed \ourdataset,
    showing that \ourdataset~provides higher-quality and much richer annotations.
    \textbf{\#Gp}: number of image groups.
    \textbf{\#Img}: number of images.
    \textbf{\#Avg}: average number of images per group.
    \textbf{IL}: whether or not instance-level annotations are provided.
    \textbf{Ceg}: whether or not category labels are provided for each group.
    \textbf{BBx}: whether or not bounding box labels are provided for each image.
    \textbf{HQ}: high-quality annotation.
  }
  \label{tab:DatasetSummary}
  \begin{tabular}{r||ccrr|cccc|r}
  \rowcolor{mygray}
  \hline\toprule
   Dataset~~~~ & Year & \#Gp & \#Img & \#Avg & IL~ & Ceg~ & BBx~ & HQ~ & Input\\
  \hline
  \hline
  \textit{MSRC}~\cite{winn2005object}   & 2005  & 8  & 240 & 30   &  & & &  & Group images\\
  \textit{iCoSeg}~\cite{batra2010icoseg}& 2010  & 38 & 643 & 17 & &  & &\checkmark & Group images\\
  \textit{Image Pair}~\cite{li2011co}   & 2011  & 105 & 210 & 2 & & $\checkmark^*$ & &  & Two images\\
  \textit{CoSal2015}~\cite{zhang2015co} & 2015  & 50  & 2,015 & 40 &  & $\checkmark^*$ & & \checkmark & Group images\\
  \textit{WICOS}~\cite{yu2018co}        & 2018  & 364  &  364  & 1 & & & & \checkmark & Single image\\
  \hline
  \emph{\textbf{CoSOD3k}}         & 2020 & 160 &3,316 & 21 & \checkmark & \checkmark & \checkmark & \checkmark & Group images \\
  \hline\toprule
  \end{tabular}
    \begin{tablenotes}
    \item * denotes coarse category rather than explicitly accurate category.
   \end{tablenotes}
  \end{threeparttable}
\end{table}

\begin{table*}[t!]
  \centering
  \scriptsize
  \renewcommand{\arraystretch}{0.95}
  \renewcommand{\tabcolsep}{0.2mm}
  \caption{Summary of 40 classic and cutting-edge CoSOD approaches.
  \textbf{Training set:} PV = PASCAL VOC07~\cite{everingham2010pascal}. CR = Coseg-Rep~\cite{dai2013cosegmentation}.
  DO = DUT-OMRON~\cite{yang2013saliency}. COS = COCO-subset.
  \textbf{Main Component:} IMC = Intra-Image Contrast.
  IGS: Intra-Group Separability.
  IGC: Intra-Group Consistency. SPL: Self-Paced Learning.
  CH: Color Histogram. GMR: Graph-based Manifold Ranking.
  CAE: Convolutional Auto Encoder. HSR: High-spatial Resolution.
  FSM: five saliency models including CBCS~\cite{fu2013cluster},
  RC~\cite{ChengPAMI15}, DCL~\cite{li2016deep},
  RFCN~\cite{wang2016saliency}, DWSI~\cite{yu2018co}.
  \textbf{SL.} = Supervision Level.
  W = Weakly-supervised. S = Supervised. U = Unsupervised.
  \textbf{Sp.:} Whether or not superpixel techniques are used.
  \textbf{Po.:} Whether or not proposal algorithms are utilized.
  \textbf{Ed.:} Whether or not edge features are explicitly used.
  \textbf{Post.:} Whether or not post-processing methods, such as,
  CRF~\cite{krahenbuhl2011efficient}, GraphCut (GCut),
  or adaptive/constant threshold (THR), are introduced.
  $\ddag$ denotes deep models.
  More details about these models can be found in recent survey papers~\cite{fan2020taking,cong2018review,zhang2018review}.
  }\label{tab:CosalModelSummary}
  \vspace{-8pt}
  \begin{tabular}{cr||rcccc|ccccc}
  \hline\toprule
  \#  & Model & Pub. & Year & \#Training & Training Set & Main Component & SL. & Sp. & Po. & Ed.& Post. \\
  \hline
  \hline
  \rowcolor{mygray}
  1&  WPL~\cite{jacobs2010cosaliency}         & UIST  & 2010 &          &  & Morphological, Translational Alignment & U   & &  &    &  \\
  2&  PCSD~\cite{chen2010preattentive}        & ICIP  & 2010 & 120,000   & 8*8 image patch & Sparse Feature~\cite{hou2009dynamic}, Filter Bank & W &  & &   &  \\
  \rowcolor{mygray}
  3&  IPCS~\cite{li2011co}                    & TIP   & 2011 &          &  & Ncut, Co-multilayer Graph  & U & \checkmark &   &  &  \\
  4&  CBCS~\cite{fu2013cluster}               & TIP   & 2013 &          &  & Contrast/Spatial/Corresponding Cue & U &&   & & \\
  \rowcolor{mygray}
  5&  MI~\cite{li2013co}                      & TMM   & 2013 &          &  & Feature/Images Pyramid, Multi-scale Voting  &  U & \checkmark  &  &  & GCut\\
  6&  CSHS~\cite{liu2013co}                   & SPL   & 2013 &          &  & Hierarchical Segmentation, Contour Map~\cite{arbelaez2010contour} & U &  &  & \checkmark &  \\
  \rowcolor{mygray}
  7&  ESMG~\cite{li2014efficient}             & SPL   & 2014 &          &  & Efficient Manifold Ranking~\cite{xu2011efficient}, OTSU~\cite{otsu1979threshold} &U&  &  & & \\
  8&  BR~\cite{cao2014co}                     & MM    & 2014 &          &  & Common/Center Cue, Global Correspondence & U & \checkmark &  & & \\
  \rowcolor{mygray}
  9&  SACS~\cite{cao2014self}                 & TIP   & 2014 &          &  & Self-adaptive Weight, Low Rank Matrix & U & \checkmark &  & &  \\
  10& DIM$^\ddag$~\cite{zhang2015cosaliency}  & TNNLS & 2015 & 1,000 + 9,963 & \tabincell{c}{ASD~\cite{achanta2009frequency} + PV} & SDAE Model~\cite{zhang2015cosaliency}, Contrast/Object Prior  & S & \checkmark & & & \\
  \rowcolor{mygray}
  11& CODW$^\ddag$~\cite{zhang2016detection}  & IJCV  & 2016 &          & ImageNet~\cite{deng2009imagenet} pre-train  & SermaNet~\cite{sermanet2013overfeat}, RBM~\cite{bengio2009learning}, IMC, IGS, IGC & W & \checkmark & \checkmark &  &  \\
  12& SP-MIL$^\ddag$~\cite{zhang2016co}       & TPAMI & 2017 & (240+643)*10\%  &  \tabincell{c}{MSRC-V1~\cite{winn2005object} + iCoSeg~\cite{batra2010icoseg}} & SPL~\cite{zhang2015self}, SVM, GIST~\cite{siva2013looking}, CNNs~\cite{chatfield2014return} & W & \checkmark & && \\
  \rowcolor{mygray}
  13& GD$^\ddag$~\cite{wei2017group}          & IJCAI & 2017 &  9,213    & MSCOCO~\cite{lin2014microsoft} & VGGNet16~\cite{simonyan2015very}, Group-wise Feature & S &  &  &  & \\
  14& MVSRCC$^\ddag$~\cite{yao2017revisiting} & TIP   & 2017 &          &  & LBP, SIFT~\cite{lowe2004distinctive}, CH, Bipartite Graph  &  & \checkmark & \checkmark &  &  \\
  \rowcolor{mygray}
  15& UMLF~\cite{han2017unified}              & TCSVT & 2017 & (240 + 2015)*50\% &  \tabincell{c}{MSRC-V1~\cite{winn2005object} + CoSal2015~\cite{zhang2016detection}} & SVM, GMR~\cite{yang2013saliency}, Metric Learning & S & \checkmark&  &  & \\
  16& DML$^\ddag$~\cite{li2018deep}           & BMVC  & 2018 & \tabincell{c}{10,000 + \\6,232 + 5,168} &  \tabincell{c}{M10K~\cite{ChengPAMI15} + THUR15K~\cite{cheng2014salientshape} + DO} & CAE, HSR, Multistage& S &&  &  & \\
  \rowcolor{mygray}
  17& DWSI~\cite{yu2018co}                    & AAAI  & 2018 &          &  & EdgeBox~\cite{zitnick2014edge}, Low-rank Matrix, CH& S &  & \checkmark &  &  \\
  18& GONet$^\ddag$~\cite{hsu2018unsupervised}& ECCV  & 2018 &          & ImageNet~\cite{deng2009imagenet} pre-train  & ResNet-50~\cite{he2016deep}, Graphical Optimization & W & \checkmark &  &  & CRF \\
  \rowcolor{mygray}
  19& COC$^\ddag$~\cite{hsu2018co}            & IJCAI & 2018 &          & ImageNet~\cite{deng2009imagenet} pre-train & ResNet-50~\cite{he2016deep}, Co-attention Loss& W  &  &\checkmark & & CRF  \\
  20& FASS$^\ddag$~\cite{zheng2018feature}    & MM    & 2018 &          & ImageNet~\cite{deng2009imagenet} pre-train  & DHS~\cite{liu2016dhsnet}/VGGNet, Graph Optimization     & W &\checkmark &  &  & \\
  \rowcolor{mygray}
  21& PJO~\cite{tsai2018image}                & TIP   & 2018 &          &  &  Energy Minimization, BoWs        &  U & \checkmark &  &  & \\
  22& SPIG$^\ddag$~\cite{jeong2018co}         & TIP   & 2018 & \tabincell{c}{10,000+210\\+2015+240}& \tabincell{c}{M10K~\cite{ChengPAMI15}+IPCS~\cite{li2011co} + \\ CoSal2015~\cite{zhang2016detection} + MSRC-V1~\cite{winn2005object}}& DeepLab, Graph Representation &  S &\checkmark  &  && \\
   \rowcolor{mygray}
  23& QGF~\cite{jera2018quality}              & TMM   & 2018 &          & ImageNet~\cite{deng2009imagenet} pre-train  & Dense Correspondence, Quality Measure & S & \checkmark &  &  &THR\\
  24& EHL$^\ddag$~\cite{song2019easy}         & NC    & 2019 & 643       & iCoSeg\cite{batra2010icoseg} & GoogLeNet~\cite{szegedy2015going}, FSM & S & \checkmark &  &  &  \\
  \rowcolor{mygray}
  25& IML$^{\ddag}$~\cite{ren2019co}          & NC    & 2019 & 3624      & \tabincell{c}{CoSal2015~\cite{zhang2016detection} + PV + CR}&   VGGNet16~\cite{simonyan2015very}     &   S   & \checkmark  &  &   &   \\
  26& DGFC$^\ddag$~\cite{wei2019deep}         & TIP   & 2019 &$>$200,000 & MSCOCO~\cite{lin2014microsoft} & VGGNet16~\cite{simonyan2015very}, Group-wise Feature & S &  \checkmark &  &  &  \\
  \rowcolor{mygray}
  27& RCANet$^\ddag$~\cite{lidetecting}       & IJCAI & 2019 &$>$200,000 & \tabincell{c}{MSCOCO~\cite{lin2014microsoft} + COS + iCoSeg~\cite{batra2010icoseg} \\ + CoSal2015~\cite{zhang2016detection} + MSRC~\cite{winn2005object}} & VGGNet16~\cite{simonyan2015very}, Recurrent Units& S &  &  &  & THR\\
  28& GS$^\ddag$~\cite{wang2019robust}        & AAAI  & 2019 &   200,000 &  COCO-SEG~\cite{wang2019robust} & VGGNet19~\cite{simonyan2015very}, Co-category Classification  & S &  &  &  &  \\
  \rowcolor{mygray}
  29& MGCNet$^{\ddag}$~\cite{jiang2019multiple}& ICME & 2019 &           &                        &     Graph Convolutional Networks~\cite{kipf2016semi}       &   S  & \checkmark  &     &   &   \\
  30& MGLCN$^{\ddag}$~\cite{jiang2019Unify}   &  MM   & 2019 &  N/A      &       N/A            &     VGGNet16, PiCANet~\cite{liu2018picanet}, Inter-/Intra-graph  &   S  &  \checkmark &  &   &   \\
  \rowcolor{mygray}
  31& HC$^{\ddag}$~\cite{li2019cosaliency}    & MM    & 2019 &  N/A      &      N/A                    &  VAE-Net~\cite{kingma2013auto}, Hierarchical Consistency    &  S    &  \checkmark  &   \checkmark  &   & CRF  \\
  32& CSMG$^\ddag$~\cite{zhang2019co}         & CVPR  & 2019 &   25,00   &  MB\cite{LiuSZTS07Learn} &   VGGNet16~\cite{simonyan2015very}, Shared Superpixel Feature         & S & \checkmark &  &  &  \\
  \rowcolor{mygray}
  33& DeepCO$^{3\ddag}$~\cite{hsu2019deepco3} & CVPR  & 2019 &   10,000  &  M10K~\cite{ChengPAMI15} &   SVFSal~\cite{zhang2017supervis} / VGGNet~\cite{simonyan2015very}, Co-peak Search & W &  & \checkmark &  &  \\
  34& GWD$^{\ddag}$~\cite{Li2019ICCV}         & ICCV  & 2019 & $>$200,000& MSCOCO~\cite{lin2014microsoft} &  VGGNet19~\cite{simonyan2015very}, RNN, Group-wise Loss &   S  &   &   &   & THR  \\
  \rowcolor{mygray}
  35 & CAFCN$^{\ddag}$~\cite{gao2020co} & TCSVT & 2020 & 200,000 & MSCOCO~\cite{lin2014microsoft} & VGGNet16~\cite{simonyan2015very}, Co-Attention, FCN & S
   & & & &\\
  36 & GSPA$^{\ddag}$~\cite{zha2020robust} & TNNLS & 2020 & 200,000 & COCO-SEG~\cite{zha2020robust} & VGGNet19~\cite{simonyan2015very}, Group Semantic, Pyramid Attention & S & & &\\
  \rowcolor{mygray}
  37 & GOMAG~\cite{jiang2020co} & TMM & 2020 & N/A & N/A & General Optimization, Adaptive Graph Learning & U & \checkmark & & &  \\
  38 & AGC$^{\ddag}$~\cite{zhang2020adaptive} & CVPR & 2020 & 200,000 & MSCOCO~\cite{lin2014microsoft} & VGGNet16~\cite{simonyan2015very}, Graph Convolution \& Clustering  & S & & & & \\
  \rowcolor{mygray}
  39 & GICD$^{\ddag}$~\cite{zhang2020Grident} & ECCV & 2020 & 8,250 & DUTS~\cite{wang2017learning} & VGGNet19~\cite{simonyan2015very}, Gradient Inducing, Attention Retaining & S & & & & \\
  \hline
  \hline
  \rowcolor{mygray}
  40& CoEG-Net$^{\ddag}$ (Ours)         &    & 2020 & 10,553 & DUTS~\cite{wang2017learning}  & VGGNet16, Co-attention Projection &   S  &   &   &  \checkmark & CRF \\

  \hline \toprule
  \end{tabular}
\end{table*}

This paper is based on and extends our previous CVPR2020 version~\cite{fan2020taking} in the following aspects.
1) We have implemented a simple but effective framework of CoSOD, which uniformly and simultaneously embeds the appearance and semantic features through a sparse convolution and a basic SOD network. Importantly, we also designed a common feature detector, which solved with Plug-and-Play.
2) We have made a lot of efforts to improve the presentations (\eg, dataset, framework, key results) and organizations of our paper.
We have added several new sections to describe our new framework about the method formulation, corresponding technical components, and further experiments (\eg, comparison with baselines, running time). Besides, several sections have been re-written to improve the readability and provide more detailed explanations about the introduction, CoSOD models, quantitative/qualitative comparisons, and discussions.
3) We build the first standard Benchmark and model zoo of CoSOD, which integrates various publicly available CoSOD datasets with uniform input/output formats (\ie, JPEG for image; PNG for GT). The gathered code of traditional or learning-based will be released soon as well.

\section{Related Work}

\subsection{CoSOD Datasets}

Currently, only a few CoSOD datasets have been proposed~\cite{winn2005object,batra2010icoseg,li2011co,cheng2014salientshape,zhang2015co,yu2018co},
as shown in \tabref{tab:DatasetSummary}.
\textit{MSRC}~\cite{winn2005object} and
\textit{Image Pair}~\cite{li2011co} are two of the earliest ones.
\textit{MSRC} was designed for recognizing
object classes from images and has spurred many interesting
ideas over the past several years. This dataset includes 8 image
groups and 240 images in total, with manually annotated pixel-level
ground-truth data.
\textit{Image Pair}, introduced by Li \etal~\cite{li2011co},
was specifically designed for image pairs and contains 210 images (105 groups)
in total.
The \textit{iCoSeg}~\cite{batra2010icoseg} dataset was released in 2010.
It is a relatively larger dataset
consisting of 38 categories with 643 images in total.
Each image group in this dataset contains 4 to 42 images,
rather than only 2 images like in the \textit{Image Pair} dataset.
The \textit{THUR15K} \cite{cheng2014salientshape} and
\textit{CoSal2015}~\cite{zhang2015co} are two large-scale publicly available datasets,
with CoSal2015 widely used for assessing co-salient object detection algorithms.
Different from the above-mentioned datasets,
the \textit{WICOS}~\cite{yu2018co} dataset
aims to detect co-salient objects from a single image,
where each image can be viewed as one group.

Although the aforementioned datasets have advanced the CoSOD task to various degrees, they are severely limited in
variety, with only dozens of groups.
On such small-scale datasets, the scalability of methods cannot be fully evaluated.
Moreover, these datasets only provide object-level labels.
None of them provide rich annotations such as bounding boxes, instances, \etc.,
which are important for progressing many vision tasks and multi-task modeling.
Especially in the current deep learning era, where models are often data-hungry.
In this work, thus, we will focus on the two relatively large-scale datasets
(\ie, \textit{iCoSeg}~\cite{batra2010icoseg} and \textit{CoSal2015}~\cite{zhang2015co})
together with the proposed challenging dataset to provide more in-depth analysis.
%


\begin{figure*}[t!]
  \centering
  \includegraphics[width=\textwidth]{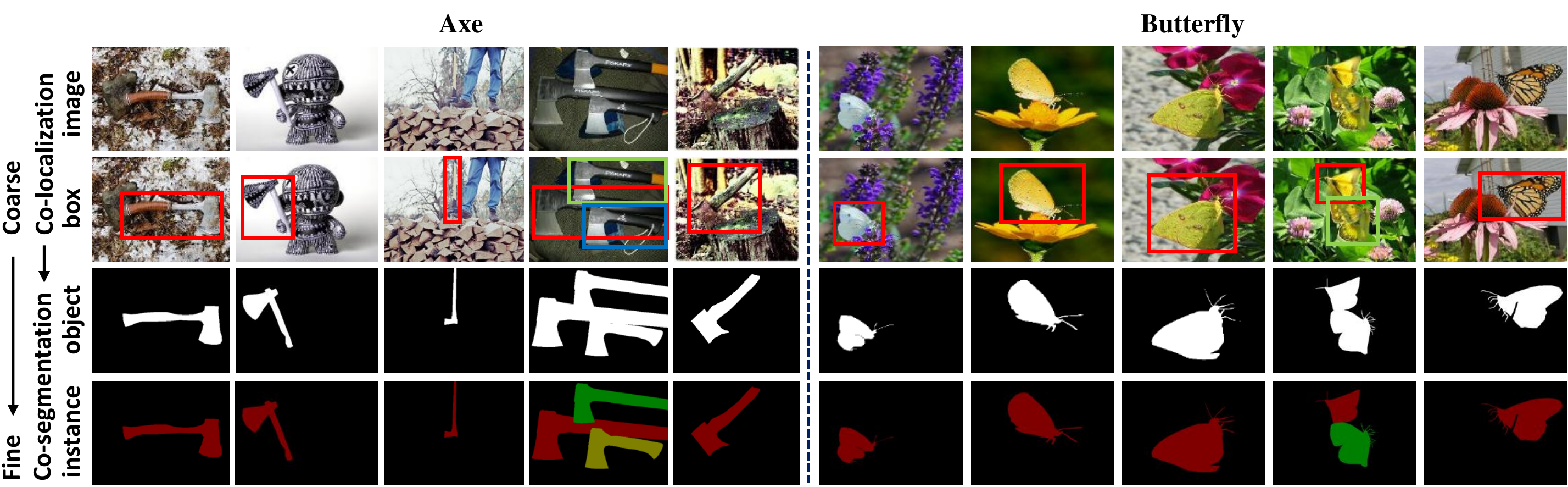}\\
  \vspace{-8pt}
  \caption{Sample images from our \ourdataset~dataset.
	  It has rich annotations, \ie, image-level categories (top),
	  bounding boxes, object-level masks, and instance-level masks.
    Our \ourdataset~will provide a solid foundation for the CoSOD task
    and can benefit a wide range of related fields, \eg, co-segmentation,
    weakly supervised localization.
  }\label{fig:CoSal3kExample}
\end{figure*}

\subsection{CoSOD Methods.}
Previous CoSOD studies~\cite{li2011co,cao2014self,han2017unified,tsai2018image}
have found that the inter-image correspondence can be effectively modeled
by segmenting the input image into several computational units
(\eg, superpixel regions \cite{zhao2018flic},
or pixel clusters~\cite{fu2013cluster}).
A similar observation can be found in recent
reviews~\cite{zhang2018review,cong2018review}.
In these approaches, heuristic characteristics (\eg,
contour~\cite{liu2013co}, color, luminance) are extracted from images,
and the high-level features are captured to express the semantic attributes in
different ways, such as through metric learning~\cite{han2017unified} or
self-adaptive weighting~\cite{cao2014self}.
Several studies have also investigated how to capture inter-image constraints
through various computational mechanisms, such as translational
alignment~\cite{jacobs2010cosaliency}, efficient manifold
ranking~\cite{li2014efficient}, and global correspondence~\cite{cao2014co}.
Some methods (\eg, PCSD~\cite{chen2010preattentive},
which only uses a filter bank technique)
do not even need to perform the correspondence matching between the two input
images, and are able to achieve CoSOD before the co-attention occurs.

%
Recently, deep learning based CoSOD models have achieved good performance
by learning co-salient object representations jointly.
For instance, Zhang~\etal~\cite{zhang2015cosaliency}
introduced a domain adaption model to transfer prior knowledge for CoSOD.
Wei \etal~\cite{wei2017group} used a group input and output to discover
the collaborative and interactive relationships between
group-wise and single-image feature representations,
in a collaborative learning framework.
Along another line, the MVSRCC~\cite{yao2017revisiting} model employs
typical features, such as SIFT, LBP, and color histograms,
as multi-view features.
In addition, several other methods~\cite{hsu2018co,jeong2018co,song2019easy,
wei2019deep,wang2019robust,zhang2019co,hsu2019deepco3}
are based on more powerful CNN models
(\eg, ResNet \cite{he2016deep}, Res2Net \cite{pami20Res2net},
GoogLeNet \cite{szegedy2015going}, and VGGNet \cite{simonyan2015very}),
achieving SOTA performances.
These deep models generally achieve better performance through either
weakly-supervised (\eg, CODW~\cite{zhang2016detection}, SP-MIL~\cite{zhang2016co},
GONet~\cite{hsu2018unsupervised}, and FASS~\cite{zheng2018feature}) or
fully supervised learning (\eg, DIM~\cite{zhang2015cosaliency} and GD~\cite{wei2017group},
and DML~\cite{li2018deep}).
There are also some concurrent works~\cite{gao2020co,zha2020robust,jiang2020co,zhang2020adaptive,zhang2020Grident} that are proposed after this submission.
A summary of the existing CoSOD models is provided in \tabref{tab:CosalModelSummary}.

\section{\ourdataset~Dataset}

\subsection{Image Collection}
We build a high-quality dataset, \ourdataset,
images of which are collected from the large-scale object recognition dataset
ILSVRC~\cite{russakovsky2015imagenet}.
There are several benefits of using ILSVRC to generate our dataset.
First, ILSVRC is gathered from \textit{Flickr} using scene-level queries
and thus it includes various object categories, diverse realistic-scenes,
and different object appearances,
and covers a large span of the major challenges in CoSOD,
providing us a solid basis for building a representative benchmark dataset
for CoSOD.
More importantly, though, the accompanying axis-aligned bounding boxes for each target
object category allow us to identify unambiguous instance-level annotations.

\subsection{Hierarchical Annotation}
Similar to \cite{mo2019partnet,fan2020Camouflage}, the data annotation is performed
in a hierarchical (coarse to fine) manner (see \figref{fig:CoSal3kExample}).


\begin{figure*}[t!]
  \centering
  \includegraphics[width=\textwidth]{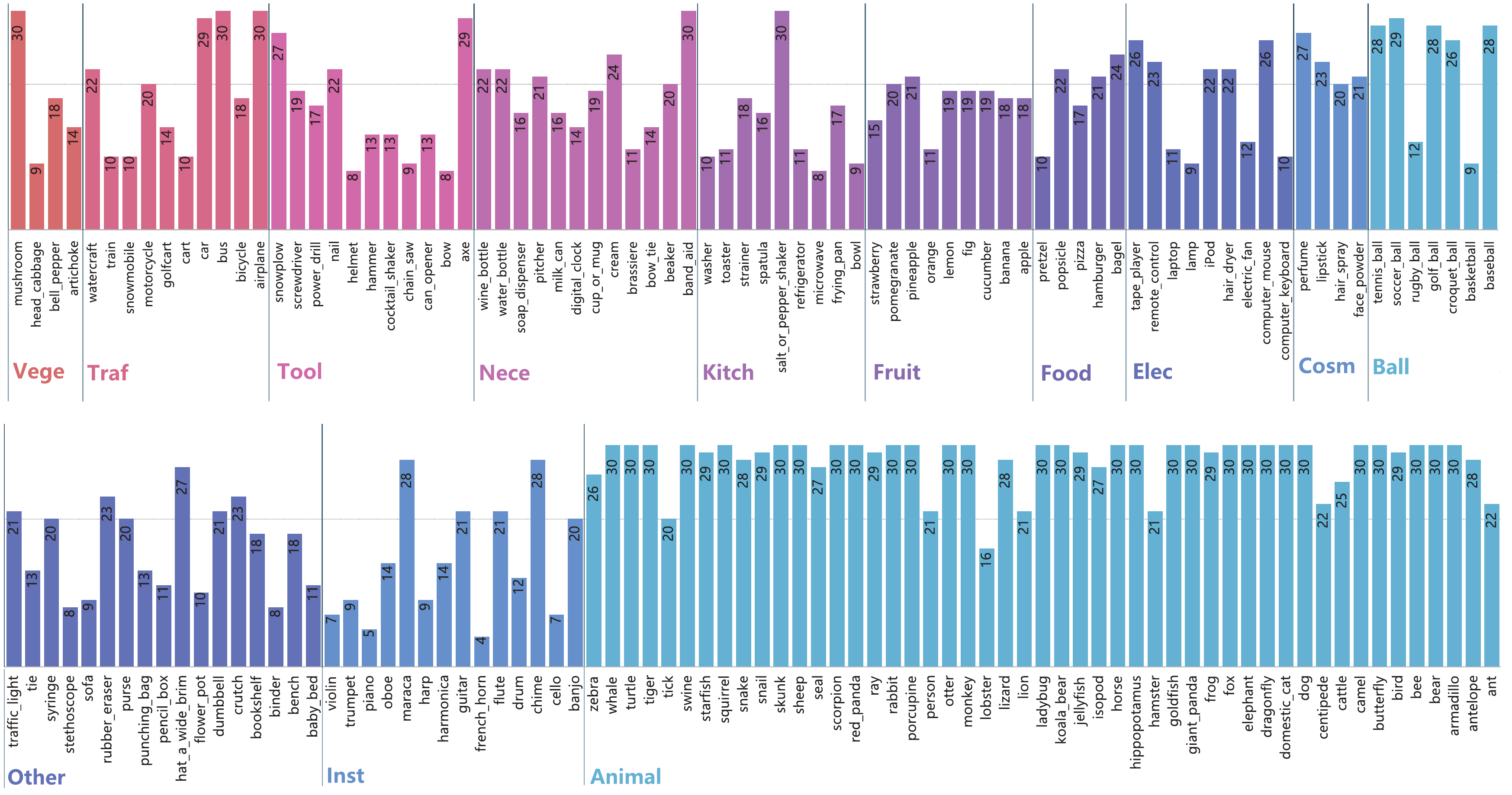}\\
  \vspace{-8pt}
  \caption{Number of images in the 160 sub-classes of our dataset.
    Best viewed on screen and zoomed-in for details.
  }\label{fig:sub-class-distribution}
\end{figure*}

\begin{figure}[t!]
  \centering
  \includegraphics[width=\columnwidth]{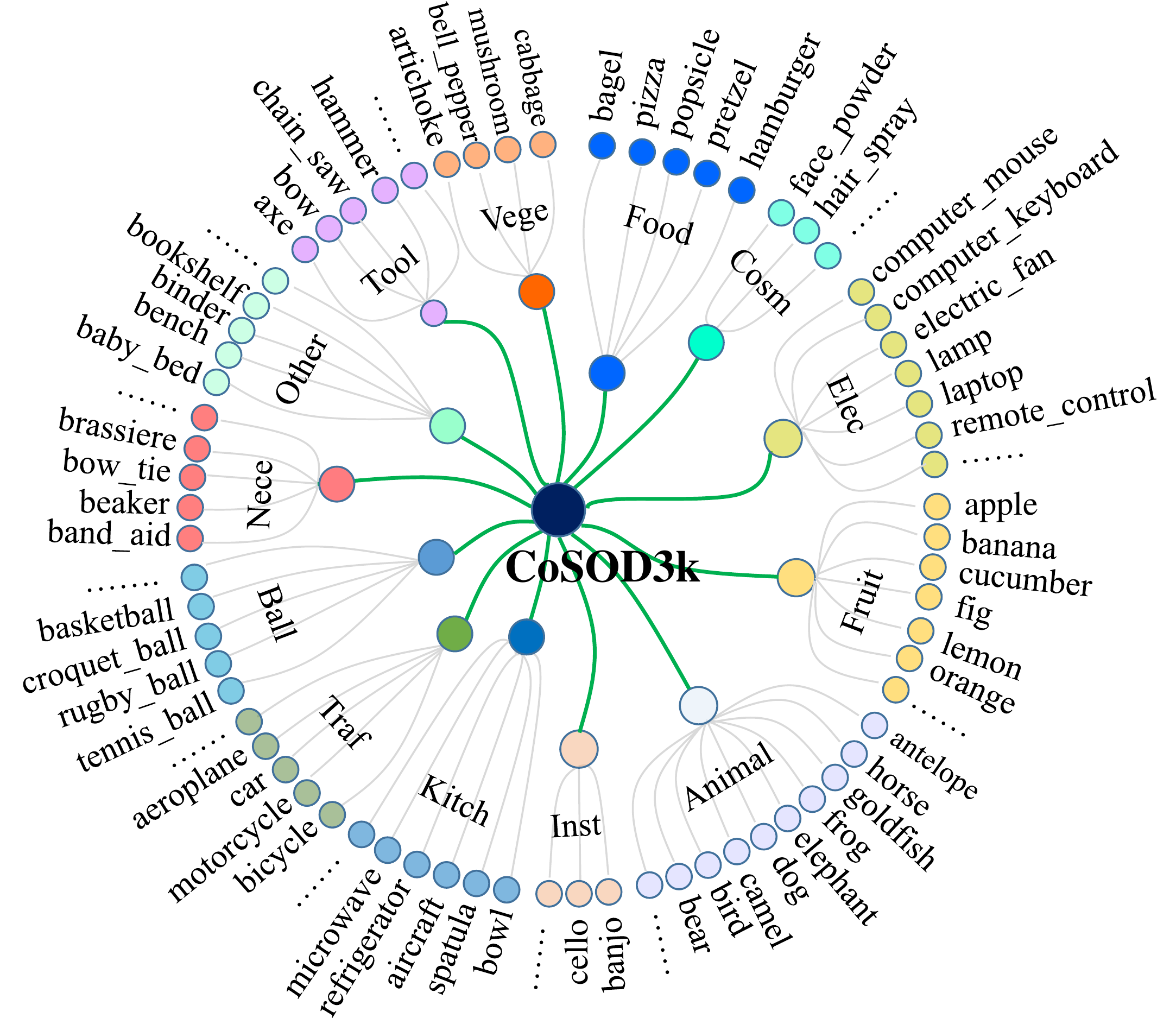}\\
  \vspace{-8pt}
  \caption{Taxonomic structure of our dataset, which contains 13 super-classes
    with 160 sub-classes.
  }\label{Taxonomy}
\end{figure}

\myPara{Category Labeling.}
We establish a hierarchical (three-level) taxonomic system
for the \ourdataset~dataset.
160 common categories (see \figref{fig:sub-class-distribution}) are selected
to generate \emph{sub-classes}
(\eg, \emph{Ant}, \emph{Fig}, \emph{Violin}, \emph{Train}, \etc.),
which are consistent with the original categories in ILSVRC.
Then, an upper-level class (\emph{middle-level}) is assigned for each \emph{sub-class}.
Finally, we integrate the upper-level classes into 13 \emph{super-classes}.
The taxonomic structure of our \ourdataset~is given in \figref{Taxonomy}.


\myPara{Bounding Box Labeling.}
The second level of annotation is bounding box labeling,
which is widely used in object detection and localization.
Although the ILSVRC dataset provides
bounding box annotations, the labeled objects
are not necessarily salient.
Following many famous SOD datasets
\cite{achanta2009frequency,alpert2007image,ChengPAMI15,
JointSalExist17,li2017instance,li2015visual,LiuSZTS07Learn,
2001iccvSOD,wang2017learning,xia2017and,yan2013hierarchical},
we ask three viewers to re-draw the bounding boxes around the object(s) in each image
that dominate their attention.
Then, we merge the bounding boxes labeled by the three viewers and have two
additional senior researchers in the CoSOD field double-check the annotations.
After that, as done in~\cite{kaufman1949discrimination},
we discard the images that contain more than six objects.
Finally, we collect 3,316 images within 160 categories.
Examples can be found in \figref{fig:CoSal3kExample}.

\begin{figure}[!t]
  \centering
  \includegraphics[width=\columnwidth]{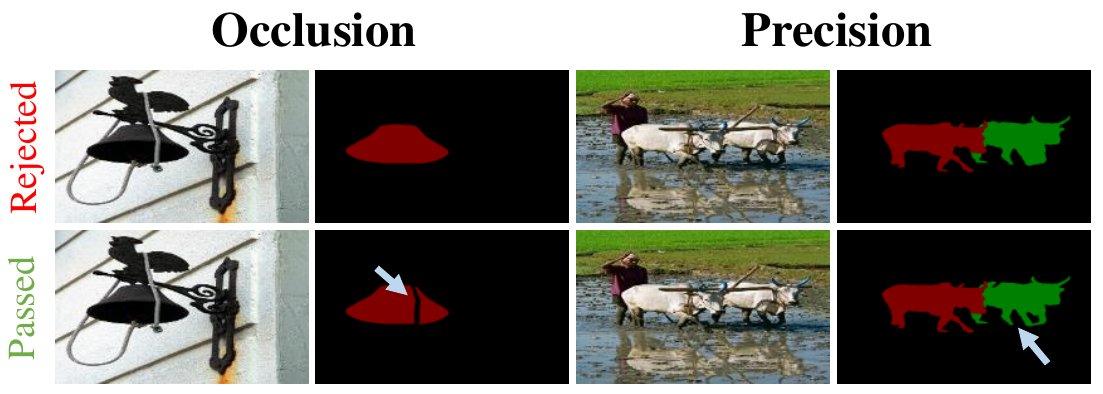} \\
  \vspace{-8pt}
	\caption{Some passed and rejected cases (\eg, occlusion, precision)
    in our \ourdataset.
  }\label{fig:QualityControl-1}
\end{figure}

\myPara{Object-/Instance-level Annotation.}
High-quality pixel-level masks are necessary for CoSOD datasets.
We hire twenty professional annotators and train them with 100 image examples.
They are then instructed to annotate the images with object- and instance-level
labels according to the previous bounding boxes.
The average annotation time per image is about 8 and 15 minutes for
object-level and instance-level labeling, respectively.
Moreover, we also have three volunteers cross-check the whole process
(more than three-fold), to ensure high-quality annotation
(see \figref{fig:QualityControl-1}).
In this way, we obtain an accurate and challenging dataset with a total of 3,316 object-level and 4,915 instance-level annotations.
Note that our final bounding box labels are refined further based on
the instance-level annotations to tighten the target.

\subsection{Dataset Features and Statistics}
To provide deeper insight into our \ourdataset, we
present several important characteristics below.

\begin{figure*}[t!]
	\centering
  \includegraphics[width=.9\textwidth]{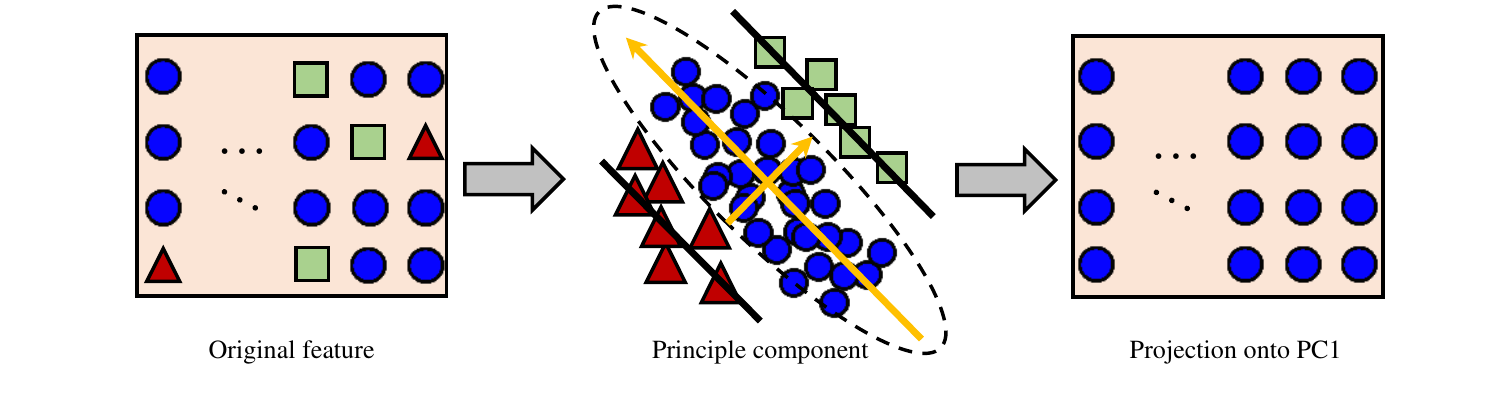} \\
  \vspace{-15pt}
  \caption{Illustration of our co-attention projection operation.
    Given the original feature representation which covers common objects (circle), noisy foregrounds (triangle) and background clutter (square),
    the co-attention projection identifies the principle
    components of common objects,
    helping to preserve the common objects while removing interference.
    By adopting our co-attention projection operation,
    we finally project the principle component and obtain the new feature representation.
    Please refer to \secref{sec:co-attention} for more details.
  }\label{fig:SC}
\end{figure*}

\begin{figure}[t!]
  \centering
  \includegraphics[width=\columnwidth]{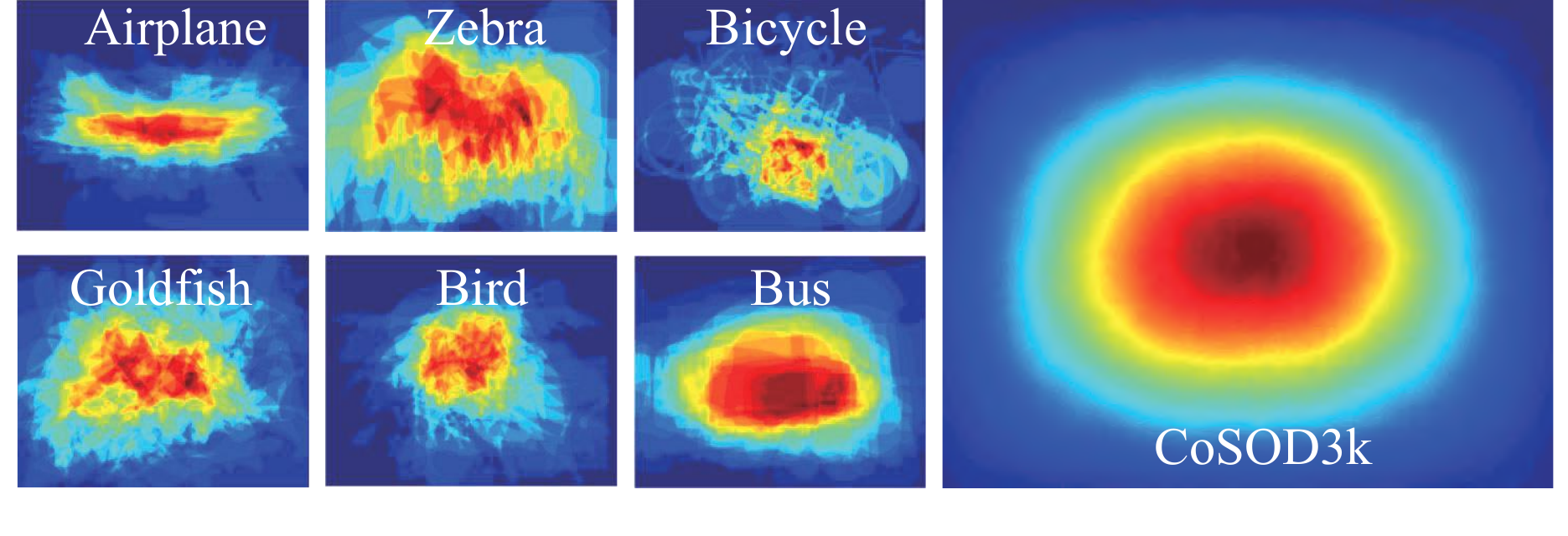} \\
  \vspace{-14pt}
  \caption{Visualization of overlap masks for mixture-specific category and overall dataset masks of our~\ourdataset.
  }\label{fig:centerbias}
\end{figure}

\myPara{Mixture-specific Category Masks.}
\figref{fig:centerbias} shows the average ground-truth masks for
individual categories and the overall dataset.
As can be observed, some categories with unique shapes
(\eg, airplane, zebra, and bicycle) present shape-biased maps,
while categories with non-rigid or convex shapes
(\eg, goldfish, bird, and bus) do not have clear shape-bias.
The overall dataset mask (the right of \figref{fig:centerbias})
tends to appear as a center-biased map without shape bias.
As is well-known, humans are usually inclined to pay more attention
to the center of a scene when taking a photo.
Thus, it is easy for a SOD model to achieve a high score when
employing a Gaussian function in its algorithm.
Due to the limitation of space, we present all 160 mixture-specific
category masks in the supplementary materials.

\begin{figure}[t!]
  \centering
  \includegraphics[width=\columnwidth]{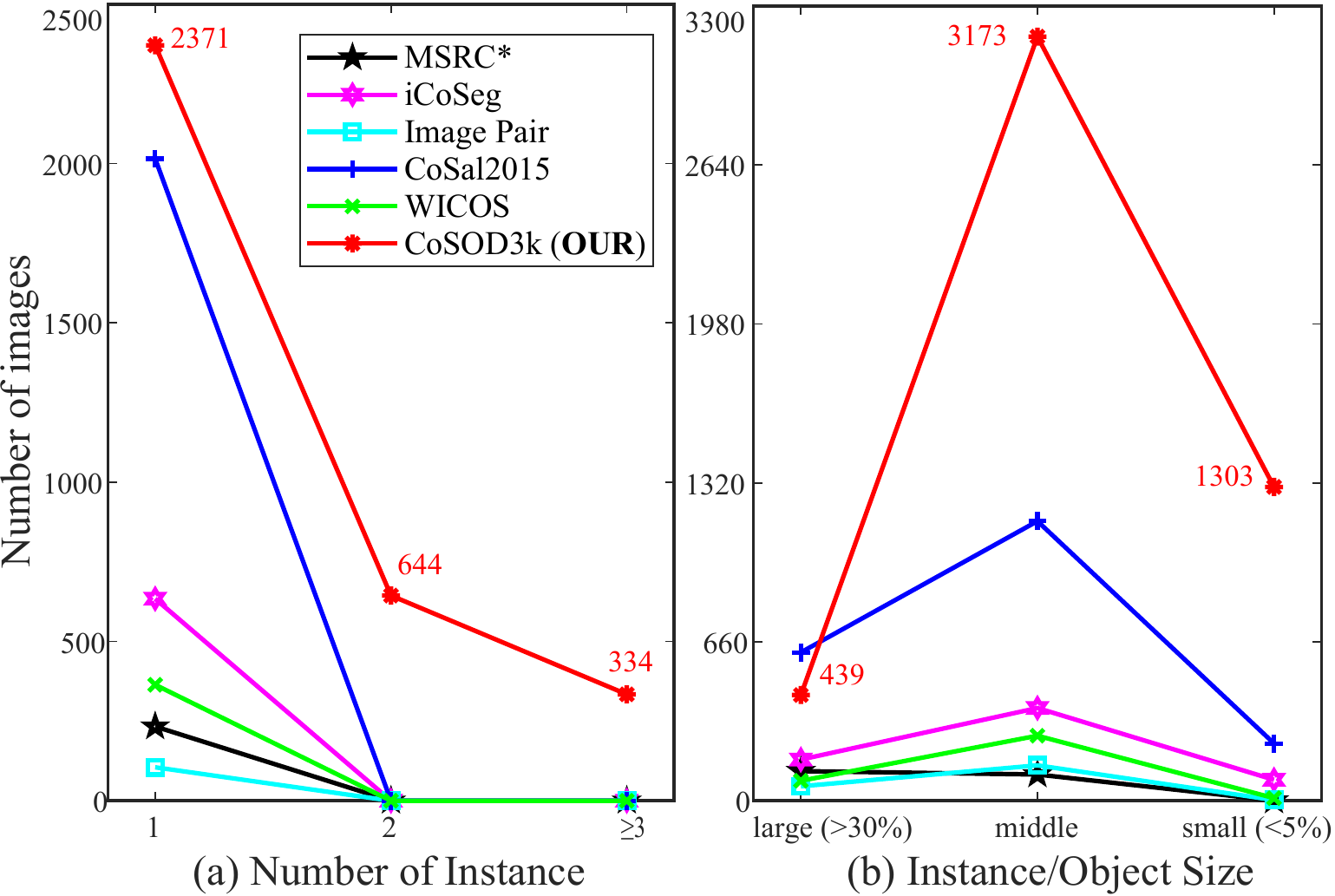} \\
  \vspace{-8pt}
  \caption{The number of images for the MSRC, iCoSeg, Image Pair, CoSal2015, WICOS, and our CoSOD3k dataset in terms of the number of instances (a) and the instance/object size (b).
  }\label{fig:statistics_of_dataset}
\end{figure}


\myPara{Sufficient Object Diversity.}
As shown in \tabref{tab:superClassPerformance} (2$^{nd}$ row) and
\figref{fig:sub-class-distribution},
our \ourdataset~covers a large variety of
super-classes including \emph{Vegetables}, \emph{Food}, \emph{Fruit},
\emph{Tool}, \emph{Necessary}, \emph{Traffic}, \emph{Cosmetic},
\emph{Ball}, \emph{Instrument}, \emph{Kitchenware}, \emph{Animal}, and
\emph{Others}, enabling a comprehensive understanding of real-world scenes.

\myPara{Number of Instances.}
Being able to parse objects into instances is critical for
humans to understand, categorize, and interact with the world.
%
To enable learning methods to gain instance-level understanding,
annotations with instance labels are in high demand.
With this in mind, in contrast to existing
CoSOD datasets, our \ourdataset~contains the multi-instance scenes
with instance-level annotations.
As illustrated in \figref{fig:statistics_of_dataset}~(a), the number of instances
(1, 2, $\geq$3) is subject to a ratio of 7:2:1.

\myPara{Size of Instances.}
The instance size is defined as the ratio of
foreground instance pixels to the total image pixels.
\figref{fig:statistics_of_dataset}~(b) shown the instance sizes of our \ourdataset~in terms of small, middle, and large instance/object.
The distributions of instance sizes are $0.02\%\!\sim\!86.5\%$
(avg.: 13.8\%), yielding a broad range.

\begin{figure*}[t!]
  \centering
  \includegraphics[width=\textwidth]{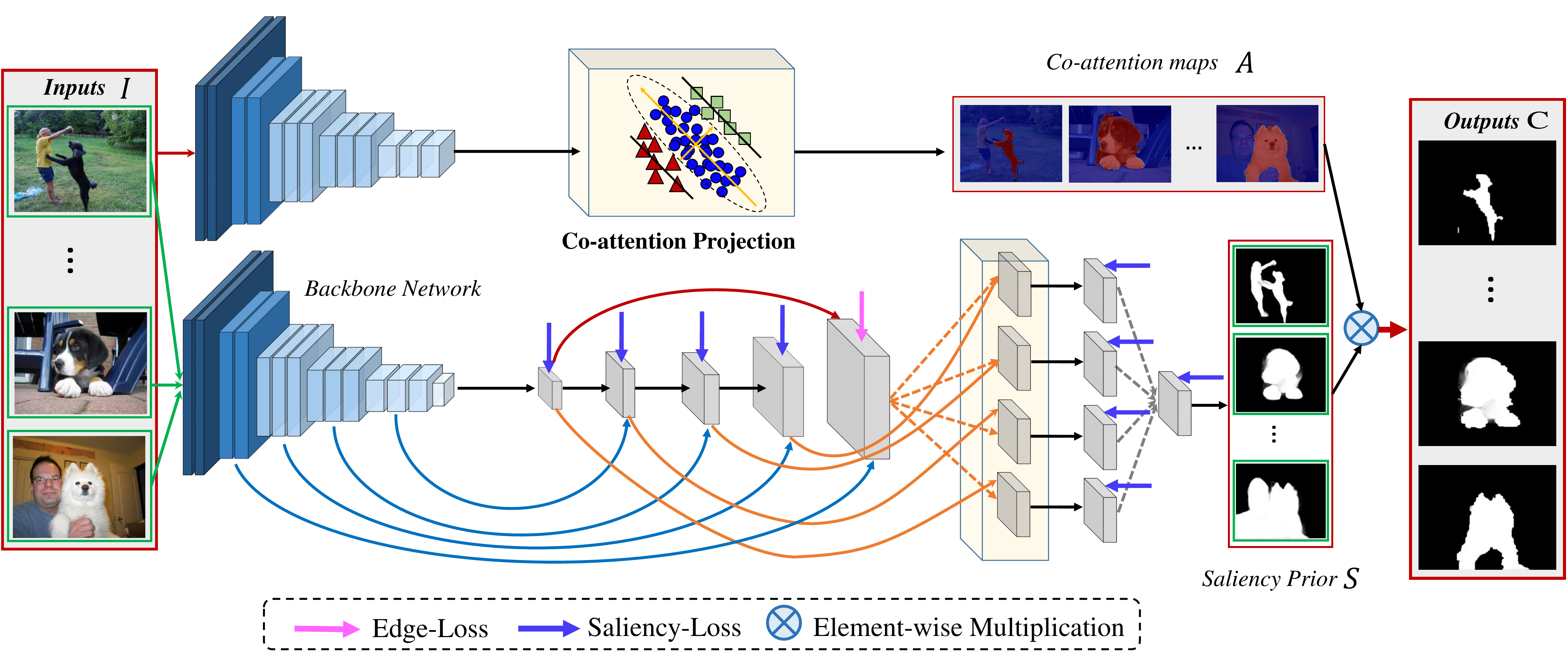} \\
  \vspace{-10pt}
  \caption{Pipeline of the proposed architecture
    which contains two separate branches.
    For a group of images $\{\mathbf{I}^n\}_{n=1}^N$
    as inputs, in the top branch,
    the extracted high-level image features are fed into the
    co-attention projection module to produce a co-attention map
    $\mathbf{A}^n$ for each input image $\mathbf{I}^n$.
    In the bottom branch, each image $\mathbf{I}^n$ is sent into the
    edge-guided saliency detection network (EGNet) \cite{zhao2019EGNet} to
    generate the saliency prior map $\mathbf{S}^n$.
    Finally, $\mathbf{A}^n$ and $\mathbf{S}^n$ are simply integrated using
    element-wise multiply to produce the optimized outputs
    $\mathbf{A}^n \otimes \mathbf{S}^n$.
    See \secref{sec:Baselines} for details.
  }\label{fig:framework}
\end{figure*}


\section{Proposed Method}\label{sec:Baselines}

In this work, we also propose a simple but effective \ourmodel~baseline
for CoSOD,
which extend \sArt SOD model EGNet \cite{zhao2019EGNet} by introducing
co-attention information in an unsupervised manner.

\subsection{Method Formulation}\label{sec:problem formulation}
For a group of $N$ associated images $\{\mathbf{I}^n\}_{n=1}^N$,
the co-saliency detection task aims at segmenting out the common
attentive foreground objects and generating optimized co-saliency maps,
which indicate common salient objects among the input images.
To predict the co-saliency masks, we present a two-branch
detection framework to respectively capture the concurrent dependencies
and salient foregrounds in a multiply independent fashion.
\figref{fig:framework} illustrates the framework of the proposed method,
which independently outputs co-attention maps $\{\mathbf{A}^n\}_{n=1}^N$
in the top branch and saliency prior maps
$\{\mathbf{S}^n\}_{n=1}^N$ in the bottom branch.
The co-attention map $\mathbf{A}^n$ and saliency prior map $\mathbf{S}^n$
are then integrated via element-wise multiply to produce the final
co-saliency prediction $\mathbf{A}^n \otimes \mathbf{S}^n$.

To obtain the saliency prior map $\mathbf{S}^n$ for an input image
$\mathbf{I}^n$,
we simply use the edge guided salient object detection method
EGNet \cite{zhao2019EGNet} to collect multi-scale saliency priors.
The EGNet is trained on large scale single image SOD dataset
DUTS~\cite{wang2017learning},
which helps to identify the salient object regions in images without
cross image information.
The real challenge then becomes how to discover co-attention map
$\mathbf{A}^n$ in an unsupervised manner,
which we present in the next subsection.


\subsection{Co-attention Projection for Co-saliency Learning}
\label{sec:co-attention}

The design of co-attention learning (see \figref{fig:SC}) is motivated by the
class activation mapping (CAM) technique proposed by
Zhou \etal \cite{zhou2016learning}.
Given an input image $\mathbf{I}^n$,
the corresponding feature \textbf{activations} $\mathbf{X}^n$
in the last convolution layer
can be easily obtained using a standard classification network
(\eg~VGGNet~\cite{simonyan2015very}).
See \tabref{tab:Notations} for more details.

Utilizing images with only keywords labeling,
the CAM technique  aims at producing a class specific attention map
$\mathbf{M}_c^n$ for each class $c$ using the \textbf{feature maps}
$\{\mathbf{X}^n_k \}$:
\begin{equation}\label{equ_cam}
  \mathbf{M}_c^n=\sum_{k=1}^{K}{\omega^c_k \mathbf{X}_k^n},
\end{equation}
where the weights $\omega^c$ could be trained using keyword level
weak supervision \cite{zhou2016learning}.
Notice that each spatial element of the class activation map $\mathbf{M}_c^n$
can be independently estimated using the weights $\omega^c$ and
the channel-wise \textbf{descriptor} in $\mathbf{X}^n$
at spatial location $(i,j)$ as
\begin{equation}\label{equ_cam_vec}
  \mathbf{M}_c^n(i,j)= (\omega^c)^\top \cdot \mathbf{x}^n(i,j).
\end{equation}
Thus the CAM \cite{zhou2016learning} technique essentially plays a linear
transformation that transforms the image features $\mathbf{x}^n(i,j)$
into class specific activation scores $\mathbf{M}_c^n(i,j)$
using the learned class specific weights $\omega^c$.

\begin{table}[t]
  \centering
  \renewcommand{\arraystretch}{1.0}
  \renewcommand{\tabcolsep}{1.5pt}
  \caption{Table of symbols, their dimensions, indices, and meaning.
  }\label{tab:Notations}
  \vspace{-8pt}
  \begin{tabular}{c|c|c|l} \hline \hline
    Symbol & Dimensions & Indices & Meaning \\ \hline
    $\mathbf{A}^n$ & $H \times W$ & $(i,j)$ &
    co-attention map of $\mathbf{I}^n$ \\ \hline
    $\mathbf{S}^n$ & $H \times W$ & $(i,j)$ &
    saliency prior map of $\mathbf{I}^n$ \\ \hline
    $\mathbf{X}^n$ & $H\times W\times K$ & $(i,j, k)$ &
    \textbf{activations} of the last conv layer \\ \hline
    $\mathbf{X}_k^n$ & $H \times W$ & $(i,j)$ &
    \textbf{feature map} of a channel in $\mathbf{X}^n$ \\ \hline
    $H$ & $1\times 1$ & scalar & spatial height \\ \hline
    $W$ & $1\times 1$ & scalar & spatial width \\ \hline
    $K$ & $1\times 1$ & scalar & number of feature channels \\ \hline
    $\mathbf{x}^n(i,j)$ & $K\times 1$ & $k$ &
    \textbf{descriptor} of $\mathbf{X}^n$ at location $(i,j)$ \\ \hline
    $\mathbf{M}_c^n$ & $H \times W$ & $(i,j)$ &
    attention map for class $c$ \\ \hline
    $\omega^c$ & $K\times 1$ & $k$ &
    channel-wise weights for class $c$ \\ \hline
    $\bar{\mathbf{x}}$ & $K\times 1$ & $k$ &
    average value of all $\mathbf{x}^n(i,j)$ \\ \hline
    $\mathbf{\hat{x}}^n(i,j)$ & $K\times 1$ & $k$ &
    $\mathbf{\hat{x}}^n(i,j) = \mathbf{x}^n(i,j) - \bar{\mathbf{x}}$ with zero mean \\ \hline
    $Cov(\mathbf{\hat{x}})$ & $K\times K$ & - &
    covariance matrix for $\{ \mathbf{\hat{x}}^n(i,j) \}$ \\ \hline
    $\xi^*$ & $K\times 1$ & k &
    first eigenvector of $Cov(\mathbf{\hat{x}})$ \\ \hline \hline
  \end{tabular}
\end{table}

\begin{table*}[t!]
  \centering
  \renewcommand{\arraystretch}{1.1}
  \renewcommand{\tabcolsep}{0.8pt}
  \footnotesize
  \caption{
    Benchmarking results of 18 leading CoSOD approaches on two
    classical~\cite{batra2010icoseg,zhang2015co}, and our \ourdataset.
   The symbol ``$\circ$'' means that the code or results are not available.
   Note that the UMLF adopts half of the images from both MSRC and CoSal2015 to train their model.
   Underline indicates the scores generated by models (\eg, SP-MIL and UMLF) that have been trained on corresponding dataset.
   See \tabref{tab:CosalModelSummary} for more training details.
  }\label{tab:BenchmarkResults}
  \begin{tabular}{lr||ccccccc|ccccccccccc|c}
  	\hline\toprule
  	\rowcolor{mygray}
  	 &               Metric &         CBCS         &          ESMG          &      RFPR       &       CSHS       &        SACS        &      CODR       &                UMLF                 &                DIM                 &               CODW                &             MIL              &           IML            &               GONet                &           SP-MIL           &            CSMG            &              CPD              &    GSPA  &  AGC      &  EGNet
  & \textbf{\ourmodel}          \\
     \rowcolor{mygray}
  	 &                      & \cite{fu2013cluster} & \cite{li2014efficient} & \cite{li2014co} & \cite{liu2013co} & \cite{cao2014self} & \cite{ye2015co} &        \cite{han2017unified}        & \cite{zhang2015cosaliency}$^\ddag$ & \cite{zhang2016detection}$^\ddag$ & \cite{zhang2015self}$^\ddag$ & \cite{ren2019co}$^\ddag$ & \cite{hsu2018unsupervised}$^\ddag$ & \cite{zhang2016co}$^\ddag$ & \cite{zhang2019co}$^\ddag$ & \cite{wu2019cascaded}$^\ddag$  &  \cite{zha2020robust}$^\ddag$ &  \cite{zhang2020adaptive}$^\ddag$ & \cite{zhao2019EGNet}$^\ddag$ & Ours$^\ddag$\\
  	 \midrule
  	 \multirow{4}{*}{\begin{sideways}iCoSeg\end{sideways}}
  	 &    $E_{\phi} \uparrow$ & .797   & .784   & .841   & .841   & .817   & .889   & .827   & .864   & .832   & .799   & .895   & .864   & \underline{.843}   & .889   & .900 &  .818 & .897 & .911  & \textbf{.912} \\
  	 & $S_{\alpha}\uparrow$ & .658   & .728   & .744   & .750   & .752   & .815   & .703   & .758   & .750   & .727    & .832   & .820   & \underline{.771}   & .821   & .861  & .784 & .821 & .875   &\textbf{.875}\\
  	 &  $F_{\beta} \uparrow$ & .705   & .685   & .771   & .765   & .770   & .823   & .761   & .797   & .782   & .741   & .846   & .832   & \underline{.794}   & .850   & .855 & .718 & .837 & .875  &\textbf{.876}\\
  	 &         $\epsilon\downarrow$ & .172   & .157   & .170   & .179   & .154   & .114   & .226   & .179   & .184   & .186   & .104   & .122   & \underline{.174}   & .106   & \textbf{.057} & .098 & .079 & .060   & .060\\
     \midrule
  	 \multirow{4}{*}{\begin{sideways}CoSal2015\end{sideways}}
  	 &    $E_{\phi} \uparrow$ & .656   & .640 & $\circ$   & .685   & .749   & .749   & \underline{.769}   & .695   & .752   & .720   & -   & .805   & $\circ$   & .842   & .841  & .855   & \textbf{.890} & .843   &.882\\
  	
  	 & $S_{\alpha}\uparrow$ & .544   & .552 & $\circ$   & .592   & .694   & .689   & \underline{.662}   & .592   & .648   & .673  & -   & .751   & $\circ$   & .774   & .814 & .797 & .823 & .818  &\textbf{.836}\\
  	 &  $F_{\beta} \uparrow$ & .532   & .476   & $\circ$   & .564   & .650   & .634   & \underline{.690}   & .580   & .667   & .620   & -   & .740   & $\circ$   & .784   & .782   & .779 & .831 & .786  & \textbf{.832}\\
  	 &         $\epsilon\downarrow$ & .233   & .247   & $\circ$   & .313   & .194   & .204   & \underline{.271}   & .312   & .274   & .210   & -   & .160   & $\circ$   & .130   & .098  & .099 & .090 & .099   & \textbf{.077}\\
  	   \midrule
    \multirow{4}{*}{\begin{sideways}CoSOD3k\end{sideways}}
    &        $E_{\phi} \uparrow$ & .637   & .635   & $\circ$   & .656   & $\circ$   & .700   & .758   & .662   & $\circ$   &
      $\circ$   & .773   & $\circ$   & $\circ$   & .804 & .791 & .800 & .823 & .793   &\textbf{.825}\\
    &        $S_{\alpha}\uparrow$ & .528  & .532    & $\circ$   & .563   & $\circ$   & .630    & .632   & .559   & $\circ$ & $\circ$   & .720   & $\circ$   & $\circ$ & .711   & .757  & .736 & .759 &  .762  & \textbf{.762}\\
     &        $F_{\beta} \uparrow$ & .466   & .418   & $\circ$   & .484   & $\circ$   & .530   & .639   & .495   & $\circ$   &
     $\circ$   & .652   & $\circ$   & $\circ$   & .709   & .699 & .682 & .729  & .702   &\textbf{.736}\\
       &         $\epsilon\downarrow$ & .228   & .239   & $\circ$   & .309   & $\circ$   & .229  & .285   & .327   & $\circ$   & $\circ$   & .164   & $\circ$   & $\circ$   & .157   & .120  & .124 & .094 & .119   & \textbf{.092}\\
  	 \hline\toprule
  \end{tabular}
\end{table*}

Unfortunately, in the co-saliency detection problem settings,
the keywords level supervision is not available.
Thus, we have to discover the weighting $\omega$ for the common objects
in an unsupervised fashion,
by revealing the internal structure of the image features.
Ideally, the unknown common object category among a group of associated images
$\{\mathbf{I}^n\}_{n=1}^N$ should corresponds to a linear projection
that results in high class activation scores in the common object regions,
while having low class activation scores in other image regions.
From another point of view,
the common object category should correspond to the linear transformation that
generates the highest variance (most informative)
in the resulting class activation maps.
Follow the idea in coarse localization task~\cite{wei2019unsupervised},
we achieve this gold by exploring the classical principle component analysis
(PCA)~\cite{pearson1901liii},
which is the simplest way of revealing the internal structure of the
data in a way that best explains the variance in the data.

Specifically, given the associated images $\{\mathbf{I}^n\}_{n=1}^N$,
with corresponding feature activations $\mathbf{X}^n$
for each image $\mathbf{I}^n$,
we aims at finding the linear transformation of $\mathbf{X}^n$
that results in the co-attention maps $\{\mathbf{A}^n \}$
with the highest variance.
This can be achieved by analyzing the co-variance matrix of the
feature descriptors $\{\mathbf{x}^n(i,j)\}$.
Let $\bar{\mathbf{x}} = \frac{1}{Z}\sum_n \sum_{i,j} \mathbf{x}^n(i,j)$,
where $Z = N \times H \times W$.
We have the zero mean version of the descriptors as
$\mathbf{\hat{x}}^n(i,j) = \mathbf{x}^n(i,j) - \bar{\mathbf{x}}$.
The covariance matrix can be denoted as
\begin{equation}\label{equ:covM}
  Cov(\mathbf{\hat{x}}) = \frac{1}{Z}\sum_{n}\sum_{i,j}
  (\mathbf{\hat{x}}^{n}(i,j)-\bar{\mathbf{x}})
  (\mathbf{\hat{x}}^{n}(i,j)-\bar{\mathbf{x}})^T.
\end{equation}
Then the expected linear projection can be established by using
the eigenvector $\xi^*$,
that corresponds to the largest eigenvalue of $Cov(\mathbf{\hat{x}})$.
Thus, the co-attention projection can be designed as a projection
that presents the features in its most informative viewpoint
\begin{equation}\label{equ:CoAtt}
  \mathbf{A}^n(i,j) = {\xi^*}^\top \cdot \mathbf{\hat{x}}^{n}(i,j).
\end{equation}

\begin{figure}[t!]
  \centering
  \includegraphics[width=\columnwidth]{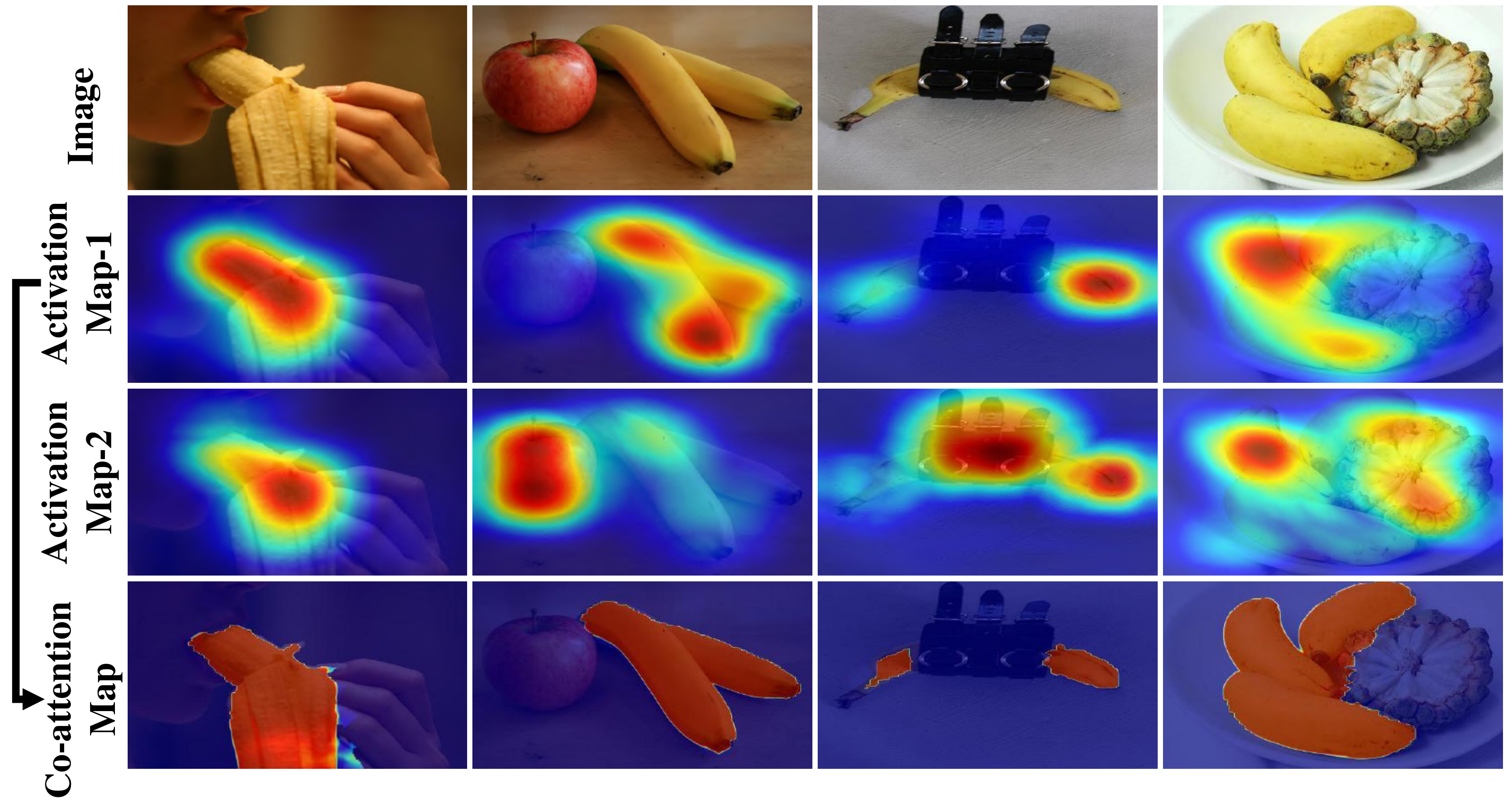} \\
  \vspace{-8pt}
  \caption{Visualization of the common activation maps (second and third row),
    using largest and second eigenvalue, and their corresponding
    post-processed (\ie, manifold ranking and DenseCRF) co-attention map $\mathbf{A}^n$ (fourth row)
    selected from the ``banana'' group of CoSal2015~\cite{zhang2015co}.
	}\label{fig:heatmap}
\end{figure}

\figref{fig:heatmap} shows some visual templates of common activation maps (second and third row) resulting from Eq.~\ref{equ:CoAtt}.
The given images contains multiple objects of diverse categories
including banana, apple, bottle and pineapple,
increasing the difficulty of differentiate correct regions,
while using the largest eigenvalue of $ Cov(\mathbf{\hat{x}})$ (second row) can sufficiently localize the common objects and
mask them out (last row) initially.

\subsection{Implementation}
The VGGNet16 network~\cite{simonyan2015very} after removing the top layer is selected as our backbone for a fair comparison.
The training process is finished in 30 epochs and the learning rate is
divided by 10 after 15 epochs.
For the edge-guided contextual saliency network,
the setting is the same with~\cite{zhao2019EGNet}.
Note that in the training stage, the loss function is the same with EGNet.
Similar to the post-processing in~\cite{hsu2018unsupervised},
we utilize the DenseCRF~\cite{krahenbuhl2011efficient} and
manifold ranking~\cite{zhou2004learning} to further refine co-attention map
$\mathbf{A}^n$ before integrating them with
the saliency prior map $\mathbf{S}^n$. The examples are shown in the third row of \figref{fig:heatmap}.

\section{Benchmark Experiments}
\label{sec_exp}

\subsection{Experimental Settings}

\myPara{Evaluation Metrics.}
To provide a comprehensive evaluation,
four widely used metrics are employed for evaluating CoSOD performance,
including maximum F-measure $F_\beta$~\cite{achanta2009frequency},
mean absolute error (MAE) $\epsilon$~\cite{Cheng_Saliency13iccv},
S-measure $S_\alpha$~\cite{fan2017structure},
and maximum E-measure $E_{\phi}$~\cite{Fan2018Enhanced}.
The complete evaluation toolbox can be found at
\url{https://github.com/DengPingFan/CoSODToolbox}.

\begin{figure*}[thp!]
	\centering
	\begin{overpic}[width=\textwidth]{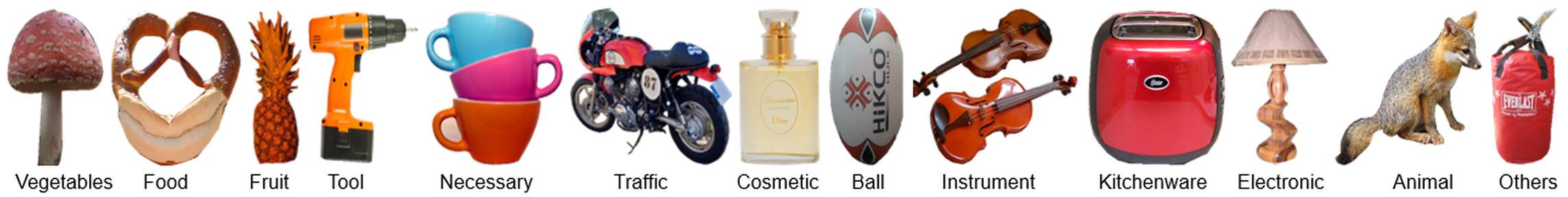}\end{overpic} \\
  \vspace{-8pt}
	\caption{ Examples of our \ourdataset. We visualize segmentation examples for representative object categories from 13 super-classes.
    }
    \label{fig:superClassExample}
\end{figure*}

\begin{table*}[t!]
  \newcommand{\R}[1]{\textcolor{red}{#1}}
  \newcommand{\G}[1]{\textcolor{green}{#1}}
  \newcommand{\B}[1]{\textcolor{blue}{#1}}
  \centering
  \renewcommand{\arraystretch}{1.1}
  \renewcommand{\tabcolsep}{6.5pt}
  \caption{Per super-class average E-measure performance $E_{\phi}$
    on our \ourdataset.
    Vege. = Vegetables, Nece. = Necessary, Traf. = Traffic, Cosm.= Cosmetic,
    Inst. = Instrument,  Kitch. =  Kitchenware, Elec. = Electronic,
    Anim. = Animal, Oth. = Others.
    ``All'' means the score on the whole dataset.
    We only evaluate the 10 state-of-the-art models with released codes.
    Note that CPD and EGNet are the top-2 SOD models on the socbenchmark
    (\url{http://dpfan.net/socbenchmark}).
  }\label{tab:superClassPerformance}
  \vspace{-8pt}
  \begin{tabular}{r||ccccccccccccc|c}
  \rowcolor{mygray}
  \hline\toprule
     & \textbf{Vege.} &\textbf{Food}  & \textbf{Fruit} &  \textbf{Tool}  & \textbf{Nece.} & \textbf{Traf.} & \textbf{Cosm.} & \textbf{Ball} & \textbf{Inst.} & \textbf{Kitch.} & \textbf{Elec.}  & \textbf{Anim.} & \textbf{Oth.} &  \textbf{All} \\
  \hline
  \#Sub-class &  4     & 5     &  9    &  11    & 12  & 10  & 4    & 7    & 14    & 9      & 9      & 49     & 17    & 160\\
  \hline
  ESMG~\cite{li2014efficient}     &   .577  &   .635  &   .735  &   .625  &   .546  &   .673  &   .633  &   .559  &   .655  &   .631  &   .629  &   .687  &   .592  &   .635  \\
  CBCS~\cite{fu2013cluster} &   .680  &   .621  &   .739  &   .617  &   .603  &   .666  &   .664  &   .619  &   .627  &   .625  &   .640  &   .672  &   .594  &   .637  \\

  CSHS~\cite{liu2013co}      &   .613  &   .591  &   .733  &   .677  &   .585  &   .691  &   .677  &   .563  &   .637  &   .651  &   .665  &   .715  &   .624  &   .656  \\

  CODR~\cite{ye2015co}     &   .682  &   .682  &   .774  &   .679  &   .634  &   .756  &   .678  &   .580  &   .671  &   .686  &   .695  &   .771  &   .638  &   .700  \\
  \hline
  DIM$^\ddag$~\cite{zhang2015cosaliency}      &   .622  &   .687  &   .773  &   .650  &   .604  &   .708  &   .633  &   .577  &   .665  &   .612  &   .641  &   .709  &   .623  &   .662  \\

  UMLF~\cite{han2017unified} & .781  & \tgb{.777} & .781  & .694  &\tbb{.779} & .836  &.714  &.668 &\tgb{.711} &   .763  &   .748  &   .810  &   .690  &   .758  \\

  IML$^\ddag$~\cite{ren2019co}      &.802 &   .725  &   .808  & \tgb{.740} &   .714  &   .867  &   .753  &.653 &\tbb{.734}&   .795  &   .729  & .855  &   .663  & .773  \\

  CPD$^\ddag$~\cite{wu2019cascaded} &\tbb{.805} &.763 & \tgb{.818} &.734 &.758 & \tgb{.894} &.763 &  .629  & .638  &\tbb{.848} & \tgb{.784} & \tbb{.892} & \tgb{.693} &.791 \\

  EGNet$^\ddag$~\cite{zhao2019EGNet}     & \trb{.833} &   .761  &.815 &\tbb{.746} & \tgb{.767} &.890 & \tbb{.769} &   .632  &   .654  &.841 &.771 & \trb{.893} & \tbb{.697} & \tgb{.793} \\

  CSMG$^\ddag$~\cite{zhang2019co}  &  .755  & \trb{.872} & \trb{.854} & .722  &   .744  & \trb{.908} & \tgb{.766} &\tbb{.778} &  .690 & \trb{.849} &\tbb{.840} & \tgb{.885} &.690 & \tbb{.804} \\
  \hline
  \hline
  \rowcolor{mygray}
  \ourmodel~(Ours)$^\ddag$ &  \tgb{.802} &  \tbb{.842} & \tbb{.840} & \trb{.811} & \trb{.790} & \tbb{.897} & \trb{.795} & \trb{.780} & \trb{.746 }& \tgb{.844} & \trb{.842} &  .881 & \trb{.739} & \trb{.825} \\

  \hline\toprule
  \end{tabular}
\end{table*}

\textbf{F-measure} $F_{\beta}$ \cite{achanta2009frequency}
evaluate the weighted harmonic mean of precision and recall.
The saliency maps have to be binarized using different threshold,
where each threshold corresponds to a binary saliency prediction.
The predicted and ground-truth binary maps are compared to get
precision and recall values.
$F_{\beta}$ is typically chosen as the F-measure score that corresponds to the
best fixed threshold for the whole dataset.

\textbf{MAE} $\epsilon$ \cite{Cheng_Saliency13iccv} is a much simple evaluation
metric that directly measures the absolute difference between the
ground-truth value and the predicted value,
without any binarization requirements.
Both F-measure and MAE evaluate the prediction in a pixel by pixel manner.

\textbf{S-measure} $S_{\alpha}$~\cite{fan2017structure} is designed
to evaluate the structural similarity between a saliency map and
the corresponding ground-truth.
It can directly evaluate the continuous saliency prediction without binarization and consider the large scale structure similarity at the same time.

\textbf{E-measure} $E_{\phi}$~\cite{Fan2018Enhanced} is a perceptual metric
that evaluates both local and global similarity between the predicted map
and ground-truth simultaneously.

%
%

\myPara{Competitors.}
In the CoSOD experiments, we evaluate/compare sixteen SOTA CoSOD models,
including seven traditional methods
\cite{fu2013cluster,li2014efficient,li2014co,liu2013co,cao2014self,ye2015co,han2017unified}
and nine deep learning models
\cite{zhang2015cosaliency,zhang2016detection,zhang2015self,ren2019co,
hsu2018unsupervised,zhang2016co,zhang2019co,wu2019cascaded,zhao2019EGNet}.
The methods were chosen based on two criteria:
(1) representative, and
(2) released code or results.

\myPara{Benchmark Protocols.}
We evaluate on two existing CoSOD datasets,
\ie, \textit{iCoSeg}~\cite{batra2010icoseg},
and \textit{CoSal2015}~\cite{zhang2015co}, and our \ourdataset.
To the best of our knowledge,
ours is the largest-scale and most comprehensive benchmark.
%
For comparison, we run the available codes directly,
either under default settings (\eg, CBCS~\cite{fu2013cluster},
ESMG~\cite{li2014efficient}, RFPR\cite{li2014co}, CSHS~\cite{liu2013co},
SACS~\cite{cao2014self}, CODR~\cite{ye2015co},
UMLF~\cite{han2017unified}, DIM~\cite{zhang2015cosaliency},
CPD~\cite{wu2019cascaded}, and EGNet~\cite{zhao2019EGNet})
or using the CoSOD maps provided by the authors (\eg, IML~\cite{ren2019co},
CODW~\cite{zhang2016detection}, GONet~\cite{hsu2018unsupervised},
SP-MIL~\cite{zhang2016co}, and CSMG~\cite{zhang2019co}). 

\subsection{Quantitative Comparisons}



\subsubsection{Performance on iCoSeg.}\label{sec:PerformanceiCoSeg}
\begin{figure*}[t!]
	\centering
	\includegraphics[width=\textwidth]{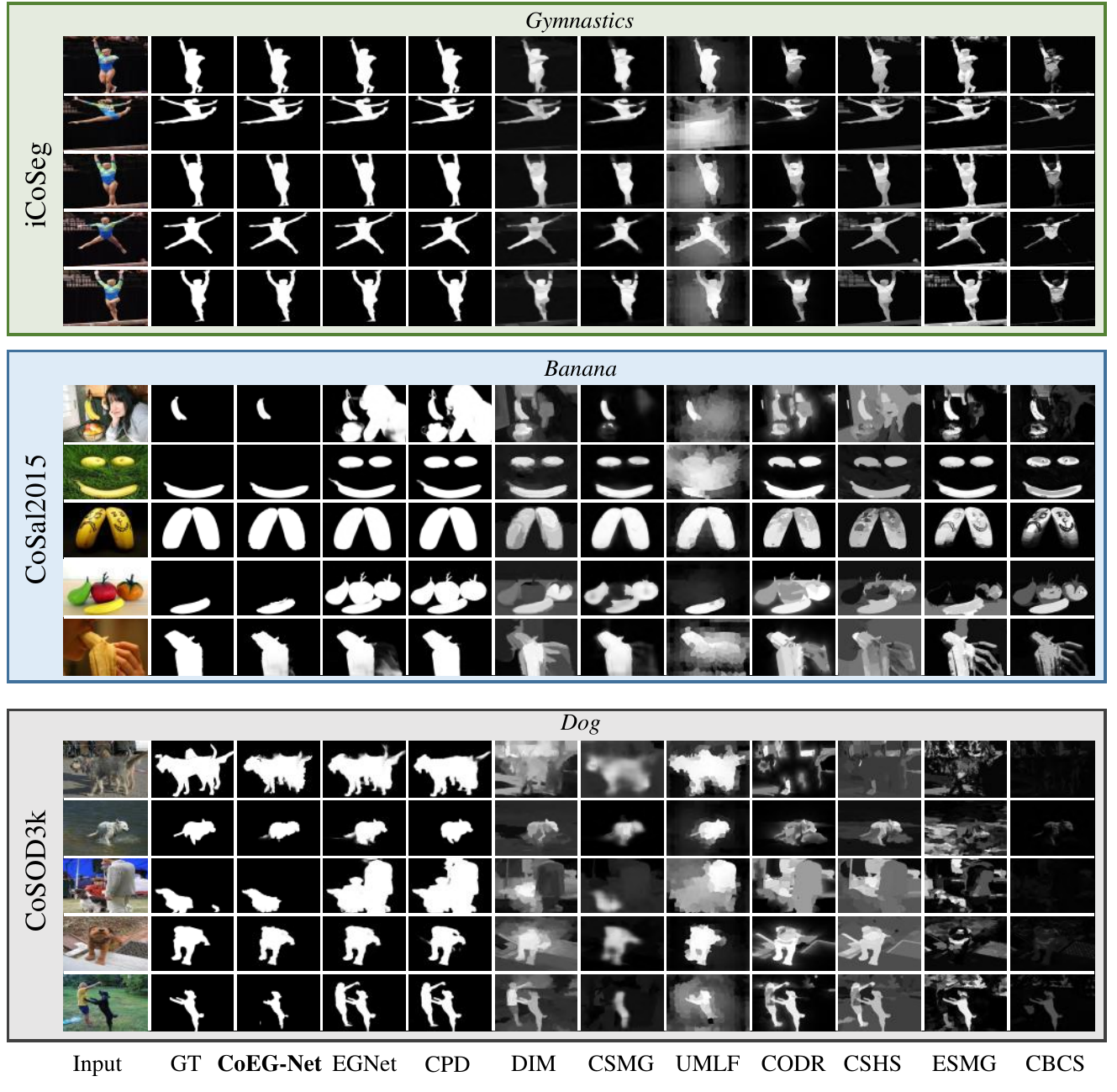} \\
	\vspace{-12pt}
	\caption{
    Qualitative examples of 10 representative models evaluated on iCoSeg~\cite{batra2010icoseg}, CoSal2015~\cite{zhang2015co}, and our \ourdataset.
  }\label{fig:compareSOTA}
\end{figure*}

The iCoSeg dataset~\cite{batra2010icoseg} was originally designed for image co-segmentation but is widely used for the CoSOD task.
Interestingly, as can be seen in \tabref{tab:BenchmarkResults}, the two SOD models
(\ie, EGNet~\cite{zhao2019EGNet} and CPD~\cite{wu2019cascaded}) achieve the state-of-the-art performances. The CoSOD methods (\eg, CODR~\cite{ye2015co}, IML~\cite{ren2019co}, and CSMG~\cite{zhang2019co}) also obtain very close performances to the top SOD models (\ie, EGNet~\cite{zhao2019EGNet} and CPD~\cite{wu2019cascaded}). Our \ourmodel~obtains the best performance in  $E_{\phi} $,  $S_{\alpha}$, and $F_{\beta} $, but the results are very close to those of the backbone, \ie, EGNet~\cite{zhao2019EGNet}.  One possible reason is that the iCoSeg dataset contains a lot of images with single objects, which can easily be detected by SOD models.
The co-salient feature is not an importance role in iCoSeg dataset. This also suggests that the iCoSeg dataset may not be suitable for evaluating CoSOD methods in the deep learning era. Some examples can be found in \figref{fig:compareSOTA}.

\subsubsection{Performance on CoSal2015.}
\tabref{tab:BenchmarkResults} shows the evaluation results on the CoSal2015 dataset~\cite{zhang2015co}. One interesting observation is that the existing salient object detection methods, \eg, EGNet~\cite{zhao2019EGNet} and CPD~\cite{wu2019cascaded}, obtain higher performances than most CoSOD methods. This implies that some top-performing salient object detection frameworks may be better-suited for extension to CoSOD tasks. The CoSOD method CSMG~\cite{zhang2019co} achieves comparable performance in $E_{\phi} $ (0.842) and $F_{\beta} $ (0.784), but worse scores in $S_{\alpha}$ (0.774) and  $\epsilon$ (0.130). This demonstrates that existing CoSOD methods cannot solve the task well. Our \ourmodel~obtains the best results, significantly outperforming both SOD and
Co-SOD baselines.




\subsubsection{Performance on CoSOD3k.}
The overall results on our \ourdataset~are presented in \tabref{tab:BenchmarkResults}. As expect, our model still achieve the best performance.
To provide deeper insight into each group, we report the performances of models on 13 super-classes in \tabref{tab:superClassPerformance}.
We observe that lower average scores are achieved on classes such as
Other (\eg, \emph{baby bed} and \emph{pencil box}), Instrument (\eg, \emph{piano}, \emph{guitar}, \emph{cello}, \etc.),
Necessary (\eg, \emph{pitcher}), Tool (\eg, \emph{axe}, \emph{nail}, \emph{chain saw}, \etc.),
and Ball (\eg, \emph{soccer}, \emph{tennis}, \etc.), which contain complex structures in
real scenes.
%
Note that almost all of the deep-based models (\eg, EGNet~\cite{zhao2019EGNet}, CPD~\cite{wu2019cascaded}, IML~\cite{ren2019co}, and CSMG~\cite{zhang2019co}) perform better than the traditional approaches (CODR~\cite{ye2015co}, CSHS~\cite{liu2013co}, CBCS~\cite{fu2013cluster}, and ESMG~\cite{li2014efficient}), demonstrating the potential advantages in utilizing deep learning techniques to address the CoSOD problem. Another interesting finding is that edge features can help provide
good boundaries for the results. For instance, the best methods from both traditional (CSHS~\cite{liu2013co}) and
deep learning models (\eg, EGNet~\cite{zhao2019EGNet}) introduce edge information to aid detection. Finally, our method \ourmodel~obtains the best performance on average, with an $E_{\phi} $ of 0.825 which is much higher than the second-best method, \ie, CSMG~\cite{zhang2019co} with 0.804. Moreover, the performances (\tabref{tab:BenchmarkResults}) of all methods are worse than on the other two datasets (\eg, iCoSeg and CoSal2015), which clearly shows that the proposed \ourdataset~dataset is challenging and leaves abundant room for further research.


\subsection{Qualitative Comparisons}
\figref{fig:compareSOTA} shows some qualitative examples on iCoSeg, CoSal2015, and our \ourdataset.
As can be seen, the SOD models, \eg, EGNet~\cite{zhao2019EGNet} and CPD~\cite{wu2019cascaded}, detect all salient objects and obtain sharp boundaries, performing better than other baselines. However, these SOD models ignore the context information.

For example, the ``banana'' group in the CoSal2015 dataset contain several other irrelevant objects, \eg, oranges, pineapples, and apples. The SOD models cannot distinguish these as being irrelevant. Another similar situation also occurs in the images in the dog group of our \ourdataset, where the humans (the third and fifth images) are detected together with the dogs.
On the other hand, the CoSOD methods, \eg, CSMG~\cite{zhang2019co} and DIM~\cite{zhang2015cosaliency}, can identify the common salient objects and remove the other objects (\eg, human). However, these CoSOD methods cannot produce accurate predicted maps, especially around object boundaries. By contrast, our \ourmodel~preserves the advantages of SOD and CoSOD methods, and obtains the best visual results in all datasets.

\begin{table}[t!]
  \caption{ Ablative studies of our model on three benchmark datasets,
    where Ours-A, Ours-P, Ours-E represent the co-salient results of Amulet,
    PiCANet, EGNet on our baseline, respectively.
  }\vspace{-8pt}
  \renewcommand{\arraystretch}{1.1}
  \renewcommand{\tabcolsep}{1.6pt}
  \footnotesize
  \begin{tabular}{c||r|cc|cc|cc} \hline \toprule
    \rowcolor{mygray}
    Datasets& Metric  & Amulet & \textbf{Ours-A} & PiCANet  & \textbf{Ours-P}
    &EGNet & \textbf{Ours-E}   \\ \hline
\multirow{4}{*}{\begin{sideways}iCoSeg\end{sideways}}
           & $E_{\phi}\uparrow$   &.877 &\textbf{.878}  &.906 &\textbf{.907}  &.911  &\textbf{.912}\\
           &$S_{\alpha}\uparrow$ &.828 &\textbf{.829}  &.869 &\textbf{.870}  &.875  &\textbf{.875}\\
           &$F_\beta \uparrow$    &.829 &\textbf{.829}  &.854 &\textbf{.854}  &.875  &\textbf{.876}\\
           &  $\epsilon\downarrow$       &.088 &\textbf{.087}  &.065 &\textbf{.064}  &.060  &\textbf{.060}\\
\hline
\multirow{4}{*}{\begin{sideways}CoSal2015\end{sideways}}
           & $E_{\phi} \uparrow$   &.772 & \textbf{.831} & .859 & \textbf{.870} &.843  &\textbf{.882}\\
           &$S_{\alpha}\uparrow$ &.719 & \textbf{.744} & .801 & \textbf{.825} &.818  &\textbf{.836}\\
           &$F_\beta \uparrow$    &.684 & \textbf{.758} & .799 & \textbf{.818} &.786  &\textbf{.832}\\
           &  $\epsilon\downarrow$       &.147 & \textbf{.125} & .090 & \textbf{.084} &.099  &\textbf{.077}\\
\hline
\multirow{4}{*}{\begin{sideways}CoSOD3k\end{sideways}}
           & $E_{\phi} \uparrow$
           & .752 & \textbf{.803}
           & .780 & \textbf{.819}
           & .793  & \textbf{.825}   \\
           &$S_{\alpha}\uparrow$
           & .685 & \textbf{.692}
           & .750 & \textbf{.758}
           & .762 & \textbf{.762}   \\
           &$F_\beta \uparrow$
           & .629 & \textbf{.700}
           & .682 & \textbf{.724}
           & .702 & \textbf{.736}    \\
           &  $\epsilon\downarrow$
           & .145 & \textbf{.122}
           & .137 & \textbf{.095}
           & .119 & \textbf{.092}   \\
\hline \toprule
\end{tabular}
\label{tab:ablation}
\end{table}

\subsection{Comparison with Baselines}
Our baseline \ourmodel~consists of a co-attention projection and a basic SOD model. In order to explore the efficiency of the co-attention projection, we (1) adopt the same training dataset (\ie, DUTS~\cite{wang2017learning}) and test datasets (\ie, iCoSeg, CoSal2015, and CoSOD3k) for three SOTA SOD models (\ie, Amulet~\cite{zhang2017amulet}, PiCANet~\cite{liu2018picanet}, and EGNet~\cite{zhao2019EGNet}); and (2) apply the same co-attention projection strategy for these models, as presented in \secref{sec:Baselines}, to conduct this experiment. \tabref{tab:ablation} shows the performances of three baselines in terms of $E_\phi $, $S_\alpha$, $F_\beta $, and $\epsilon$ metrics. Based on the results, we observe that: (i) On the relatively simple iCoSeg dataset, our baselines (\ie, Ours-A/-P/-E) slightly improve upon the backbone models (\ie, Amulet, PiCANet, and EGNet). We note that because this dataset contains a large number of single objects with similar appearances (\figref{fig:compareSOTA}) in each group, only using a SOD model can achieve very high performance.
This conclusion is consistent with the analysis in \secref{sec:PerformanceiCoSeg}; (ii) On the classical CoSal2015 dataset, our baselines are consistently better than the backbones in terms of all four metrics. It is worth noting that, for this more complex dataset, we still obtain a 2.5\%, 1.4\%, and 1.8\% $S_\alpha$ score improvement; (iii) For the proposed and most challenging dataset \ourdataset , we find that the improvement is still significant (\eg, 7.1\% $F_\beta $ score for Amulet). To further analyze the improvement, we also provide the 160 sub-class performances in the \supp{supplementary materials}. We observe that, for objects in the common super-class (\ie, `Ball') such as ``rugby\_ball'' and ``soccer\_ball'', we achieve 23.5\% and 23.9\% $F_\beta $ improvements.
We attribute this to the co-attention projection operation being able to automatically learn mutual-features, which are crucial for overcoming challenging ambiguities.

\begin{table}[t!]
  \caption{Average running time of ten SOTA models.}\label{tab:Runtime}
  \vspace{-8pt}
  \centering
  \renewcommand{\arraystretch}{1.0}
  \renewcommand{\tabcolsep}{1.8pt}
  \footnotesize
  \begin{tabular}{c||c|c|c|c} \hline \toprule
  \rowcolor{mygray}
  Models
    & CBCS~\cite{fu2013cluster}
    & ESMG~\cite{li2014efficient}
    & CSHS~\cite{liu2013co}
    & CODR~\cite{ye2015co} \\ \hline
  Time (seconds) & 0.3 & 1.2 & 102 & 35 \\ \hline
  Language & Matlab &  Matlab &  Matlab &  Matlab\\ \hline
  \rowcolor{mygray}
  Models
    & UMLF~\cite{han2017unified}
    & DIM$^\ddag$~\cite{zhang2015cosaliency}
    & CSMG$^\ddag$~\cite{zhang2019co}
    & CPD$^\ddag$~\cite{wu2019cascaded} \\ \hline
  Time (seconds) & 87 & 25 & 3.2 & 0.016\\ \hline
  Language & Matlab &  Matlab &  Caffe &  PyTorch\\ \hline
  \rowcolor{mygray}
  Models & EGNet$^\ddag$~\cite{zhao2019EGNet} & Ours$^\ddag$ & & \\ \hline
  Time (seconds) &  0.034 & 2.3 & & \\ \hline
  Language & PyTorch &  PyTorch & & \\
\hline \toprule
\end{tabular}
\end{table}

\subsection{Running Time}
Our \ourmodel~is implemented in PyTorch and Caffe with an RTX 2080Ti GPU for acceleration. For traditional algorithms (CBCS~\cite{fu2013cluster}, ESMG~\cite{li2014efficient}, CSHS~\cite{liu2013co}, CODR~\cite{ye2015co}, and UMLF~\cite{han2017unified}), the comparison experiments are executed on a laptop with Inter(R) Core(TM) i7-2600 CPU @3.4GHz. The remaining deep learning models (DIM~\cite{zhang2015cosaliency}, CSMG~\cite{zhang2019co}, CPD~\cite{wu2019cascaded}, and EGNet~\cite{zhao2019EGNet}) are tested on
a workstation with Intel(R) Core(TM) i7-8700K CPU @3.70GHz and an RTX 2080Ti GPU. As shown in \tabref{tab:Runtime}, among the top-3 CoSOD models, \ie, the proposed \ourmodel, CSMG~\cite{zhang2019co}, and UMLF~\cite{han2017unified}, evaluated in terms of $E_\phi $ measure on the proposed \ourdataset, our model achieves the fastest inference time. In addition, compared with the top-2 fastest CoSOD models (\ie, CBCS~\cite{fu2013cluster} and ESMG~\cite{zhang2019co}), although the proposed model has a longer test time, it obtains a significantly improved  $S_\alpha$ measure. This partially suggests that our framework is not only efficient but also effective for the CoSOD task. However, compared to two recently released state-of-the-art models, CPD~\cite{wu2019cascaded} and EGNet~\cite{zhao2019EGNet}, there is still large room for improvement in running time.

\section{Discussion and future directions}
From the evaluation, we observe that, in most cases, the current SOD methods (\eg, EGNet~\cite{zhao2019EGNet} and CPD~\cite{wu2019cascaded}) can obtain very competitive or even better performances than the CoSOD methods (\eg, CSMG~\cite{zhang2019co} and SP-MIL~\cite{zhang2016co}).
However, this does not necessarily mean that the current datasets are not complex enough or  using the SOD methods directly can obtain the good performances---the performances of the SOD methods on the CoSOD datasets are actually lower than
those on the SOD datasets. For example, EGNet achieves 0.937 and 0.943 $F_\beta $ scores on the HKU-IS dataset~\cite{li2015visual} and ECSSD dataset~\cite{yan2013hierarchical}, respectively. However, it only obtains 0.786 and 0.702 $F_\beta $ scores on the CoSal2015 and CoSOD3k datasets, respectively. Consequently, the evaluation results reveal that many problems in CoSOD are still under-studied and this makes the existing CoSOD models less effective. In this section, we discuss four important issues (\ie,
scalability, stability, compatibility, and metrics) that have not been fully addressed by the existing co-salient object detection methods and should be studied in the future. Finally, we discuss the weakness of the the proposed \ourmodel~framework.

\myPara{Scalability.}
The scalability is one of the most important issues that needs to be considered when designing CoSOD algorithms. Specifically, it indicates the capability of a CoSOD model of handling large-scale image scenes. As we know, one key property of CoSOD is that the model needs to consider multiple images from each group. However, in reality, an image group may contain numerous related images. Under this circumstance,
methods that do not consider scalability would have huge computational costs and take a very long time to run, making them unacceptable in practice (\eg, CSHS-102s and UMLF-87s).
Thus, how to address the scalability issue, or how to reduce the computational complexity caused by the number of images contained in an image group, becomes a key problem in this field, especially when applying CoSOD methods for real-world applications.

\myPara{Stability.}
Another important issue is the stability of model. When dealing with image groups containing multiple images, some existing methods (\eg, HCNco~\cite{lou2017hierarchical}, PCSD~\cite{chen2010preattentive}, and IPCS~\cite{li2011co}) divide the image group into image pairs or image sub-groups (\eg, GD~\cite{wei2017group}). Another school of methods adopt the RNN-based model
(\eg, GWD~\cite{Li2019ICCV}), which involves assigning an order to the input images. These strategies all make the overall training process unstable as there is no  principle way of dividing  image groups or assigning input order to related images. In other words, when generating image sub-groups or assigning the input orders following different strategies, the learning procedure produces different co-saliency detectors, and the test results are also unstable. Consequently, this not only brings difficulty for evaluating the performance of the learned co-saliency detectors but also influences the application of the co-salient object detection.

\myPara{Compatibility.}
Introducing SOD in CoSOD is a direct yet effective strategy for building  CoSOD framework as the single image saliency can conduce to the co-saliency pattern identification.
However, most existing CoSOD works only utilize the results or features of the SOD models as useful information cues.
The proposed \ourmodel~baseline still follows this two-stage framework that spends more inference time than the single SOD model. Although as a preliminary attempt, we have also achieved the best performance among the existing CoSOD models.
From this point of view, one further direction for leveraging the SOD technique is to deeply
combine a CNN-based SOD network with a CoSOD model to
build an end-to-end trainable framework for detecting CoSOD directly.
To achieve this goal, one needs to consider the compatibility of the
CoSOD framework, making it convenient for integrating
the existing SOD techniques.

\myPara{Metrics.}
Current evaluation metrics for CoSOD are designed in terms of SOD, \ie, they calculate the mean of the SOD scores on each group directly. In contrast to SOD,  CoSOD involves relationship information between co-salient objects of different images, which is more important for CoSOD evaluation.
For example, current CoSOD metrics assume taht the target objects have similar sizes in all images. As the objects actually have different sizes in different images, these metrics ($S_\alpha, E_\phi , F_\beta , \epsilon$ in Sec.~\ref{sec_exp}) would likely be inclined to detecting large objects. Moreover, the current CoSOD metrics are based towards detecting  objects in a single image, rather than  identifying co-occurring objects across multiple images. Thus, how to design suitable metrics for CoSOD is an open issue.

\begin{figure}[!t]
  \centering
  \includegraphics[width=\columnwidth]{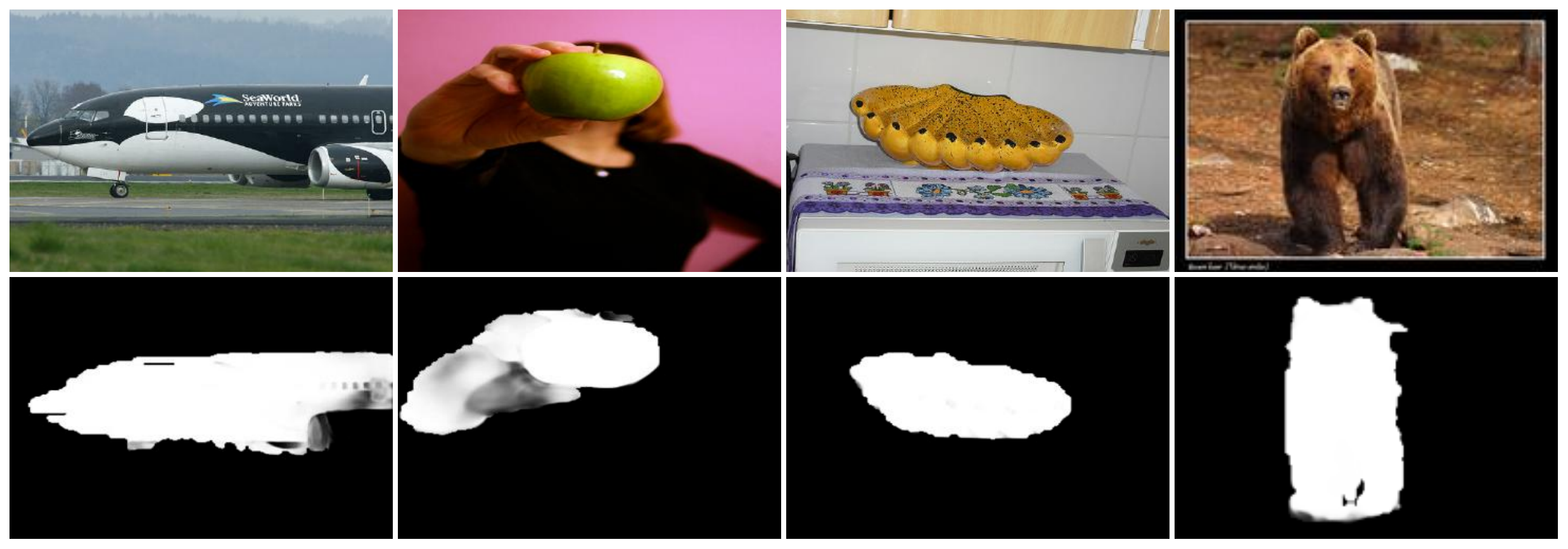} \\
  \vspace{-10pt}
  \caption{Some challenge cases for our \ourmodel.}
  \label{fig:weakness}
\end{figure}

\myPara{Weakness.}
Compared with the end-to-end CoSOD detection frameworks that output binary predictions with smoothed fine-structures, the prediction results of \ourmodel~suffers from coarse boundaries, indicating that \ourmodel~cannot finely preserve detailed shape information for the co-salient objects. Some failed detection cases are shown in Fig.~\ref{fig:weakness}.

\begin{figure}[!t]
  \centering
  \includegraphics[width=\columnwidth]{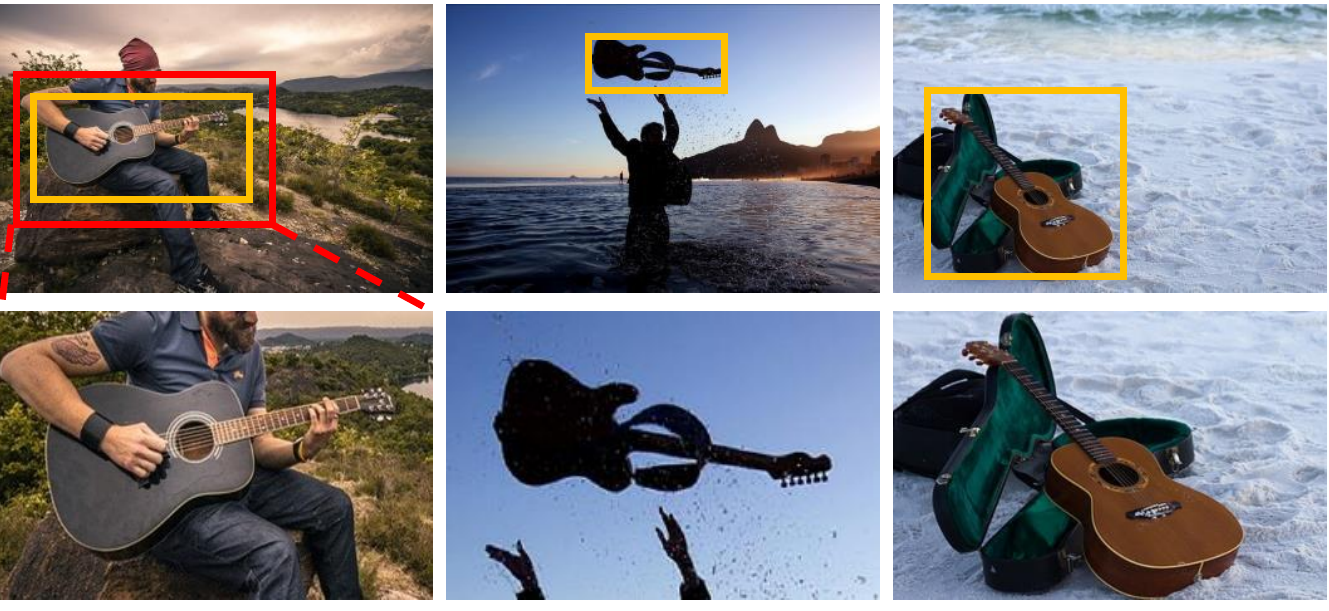} \\
  \vspace{-10pt}
  \caption{Collection-Aware Crops.}
  \label{fig:Crops}
\end{figure}

\myPara{Potential Applications.}
In this part, we discuss two  potential new applications that could benefit from the high-quality CoSOD models. For more CoSOD applications, please refer to the related survey in~\cite{zhang2018review,cong2018review}.

\textit{Collection-Aware Crops.} This application is derived from Jacob \etal's~work~\cite{jacobs2010cosaliency}. It studies where people look when comparing images and triggers the seminal works on the CoSOD task. Sharing the same spirit, we show a more general potential application which is not limited to image pairs. As an example, when dealing with the automatic thumb-nailing task as in \figref{fig:Crops}, we first obtain the yellow bounding box from the saliency maps generated by our \ourmodel.
After that, an enlarged ($\sim$60 pixels) red box\footnote{Note that we keep the original width of the yellow box when the enlarged red box touched the boundary of the image.} is used to identify the crop regions automatically. To obtain high-quality crops from the first row, we can also introduce existing SOTA super-resolution techniques~\cite{bahat2020explorable,lugmayr2020srflow} to further improve the visualization results.

\textit{Object Co-Localization.} As shown by DeepCO$^3$~\cite{hsu2019deepco3}, the co-saliency detection results will provide the class-agnostic attention cues for the object co-location task.
Introducing our \ourmodel~to existing commerce application will be a possible solution to improve the performance in this field.

\section{Conclusion}
In this paper, we have presented a comprehensive investigation on the co-salient object detection (CoSOD) task. After identifying the serious data bias in current dataset, which assume that each image group contains salient object(s) of similar visual appearance, we built a new high-quality dataset, named \ourdataset, containing co-salient object(s) that are similar at a semantic or conceptual level.
Notably, \ourdataset~is the most challenging CoSOD dataset so far, containing 160 groups and total of 3,316 images labeled with category, bounding box, object-level, and instance-level annotations.
%
Our \ourdataset~dataset makes a significant leap in terms of diversity, difficulty and scalability, benefiting several related vision tasks, \eg, co-segmentation, weakly supervised localization, and instance-level detection, and their future development.

To creat an effective co-salient object detector, we integrated existing SOD techniques to build a unified, trainable CoSOD framework called \ourmodel. Specifically, we augmented our prior model EGNet with a co-attention projection strategy to enable efficient common information learning, improving the scalability and stability of the co-salient object detection framework.

Besides, this paper has also provided a comprehensive study by summarizing 40 cutting-edge algorithms, benchmarking 18 of them over two classical datasets, as well as the proposed \ourdataset.
By evaluating recent SOD and CoSOD methods, this paper demonstrated that the SOD methods are surprisingly better. This is an interesting finding that can guide further investigation into better CoSOD algorithms.
We hope the studies presented in this work will give a strong boost to the growth of the CoSOD community.
In the future, we plan to increase the dataset scale to spark more novel ideas.

\section*{Acknowledgment}
We also thank professor Kaihua Zhang from Nanjing University of Information Science \& Technology for insightful feedback.
This research was supported by NSFC (61922046),
S\&T innovation project from Chinese Ministry of Education,
and Tianjin Natural Science Foundation (18ZXZNGX00110).

{\small
\bibliographystyle{IEEEtran}
\bibliography{CoSOD}
}

\vfill


\end{document}